%% file: main.tex
\documentclass{article}
\usepackage{meld_arxiv,times}
\usepackage{mathptmx}
\DeclareMathAlphabet{\mathcal}{OMS}{cmsy}{m}{n}  % mathptmx re-points \mathcal at rsfs swash; keep the conventional calligraphic

\usepackage[utf8]{inputenc}
\usepackage[T1]{fontenc}
\DeclareUnicodeCharacter{0430}{a}
\DeclareUnicodeCharacter{043E}{o}
\DeclareUnicodeCharacter{0435}{e}
\DeclareUnicodeCharacter{0441}{c}
\DeclareUnicodeCharacter{0440}{p}
\DeclareUnicodeCharacter{0445}{x}
\DeclareUnicodeCharacter{0443}{y}
\DeclareUnicodeCharacter{0456}{i}
\DeclareUnicodeCharacter{0410}{A}
\DeclareUnicodeCharacter{0412}{B}
\DeclareUnicodeCharacter{0421}{C}
\DeclareUnicodeCharacter{0415}{E}
\DeclareUnicodeCharacter{041D}{H}
\DeclareUnicodeCharacter{041A}{K}
\DeclareUnicodeCharacter{041C}{M}
\DeclareUnicodeCharacter{041E}{O}
\DeclareUnicodeCharacter{0420}{P}
\DeclareUnicodeCharacter{0422}{T}
\DeclareUnicodeCharacter{0425}{X}
\usepackage{hyperref}
\usepackage{url}
\usepackage{array}
\usepackage{booktabs}
\usepackage{longtable}
\usepackage{amsfonts}
\usepackage{amsmath,amssymb}
\usepackage{amsthm}
\usepackage{nicefrac}
\usepackage{microtype}
\usepackage{xcolor}
\usepackage{graphicx}
\usepackage{multirow}
\usepackage{caption}
\usepackage{subcaption}
\usepackage{float}
\usepackage{wrapfig}
\usepackage{pifont}

\newcommand{\xmark}{\ding{55}}
\usepackage{tikz}
\usetikzlibrary{arrows.meta, positioning, decorations.pathreplacing, calc, fit}
\newcommand{\tableci}[2]{\mbox{#1$\pm$#2}}

\newtheorem{proposition}{Proposition}

\newcommand{\LSE}{\operatorname{LSE}}
\newcommand{\Sai}{S_{\mathrm{AI}}}
\newcommand{\Sfam}{S_{\mathrm{fam}}}
\newcommand{\Senergy}{S_{\mathrm{energy}}}

\title{MELD: Multi-Task Equilibrated Learning Detector for AI-Generated Text}

\author{Chenjun Li\textsuperscript{1}\quad Cheng Wan\textsuperscript{1}\quad
Haomiao Chen\textsuperscript{1}\quad Johannes C. Paetzold\textsuperscript{1}\\[5pt]
\normalfont\normalsize\textsuperscript{1}Cornell University, Ithaca, NY 14853, USA}
\hypersetup{
  pdftitle={MELD: Multi-Task Equilibrated Learning Detector for AI-Generated Text},
  pdfauthor={Chenjun Li, Cheng Wan, Haomiao Chen, Johannes C. Paetzold},
  pdfsubject={AI-generated text detection},
  hidelinks
}

\begin{document}
\maketitle

\begin{abstract} Large language models are widely used in everyday writing, making reliable AI-generated text detection crucial for academic integrity, content moderation, and provenance tracking. Yet high AUROC on clean, in-distribution benchmarks is not sufficient. Practical detectors must resist adversarial rewrites, generalize to unseen generators and writing domains, and maintain low false-positive rates (FPR). 
A pooled AI-versus-human objective does not explicitly require the model to distinguish among generator families, so it may fail to learn the generator-specific structure needed for generalization and attribution. We introduce MELD (Multi-Task Equilibrated Learning Detector), which instead trains on class-balanced, family-specific AI-versus-human tasks while sharing a common representation of human writing. 
MELD produces detection, generator-family, rewrite-task, and token-level predictions in a single forward pass. Format normalization further makes its scores invariant to the modeled reformatting operations. MELD ranks first among open-source submissions in the public RAID leaderboard and matches or exceeds supervised baselines on five of six held-out evaluation pools. To evaluate transfer to unseen generators, we introduce MELD-eval, a held-out test pool built from four frontier chat models. Without further fine-tuning, MELD achieves 99.7\% TPR at 1\% FPR on MELD-eval and 98\% TPR at the same FPR on a held-out generator whose family is absent from training.
Finally, in a case study of 5.9 million scientific texts from 2016--2026, MELD's prediction scores remain stable through 2022 and increase from 2023 onward, coinciding with the widespread adoption of LLM-based writing tools. The model, MELD-eval pool, source code, and live demo are available.
\par\vspace{8pt}
\begin{center}
\meldbutton{https://anonymous.4open.science/r/MELD-4D74/README.md}{\faGithub}{Code}\hspace{8pt}%
\meldbutton{https://huggingface.co/anon-review-meld-2026/meld}{\faCube}{Model \& Data}\hspace{8pt}%
\meldbutton{https://anon-review-meld-2026-meld.static.hf.space/index.html}{\faPlayCircle}{Live Demo}
\end{center}
\end{abstract}
\begin{figure}[t]
\centering
\includegraphics[width=1.0\textwidth]{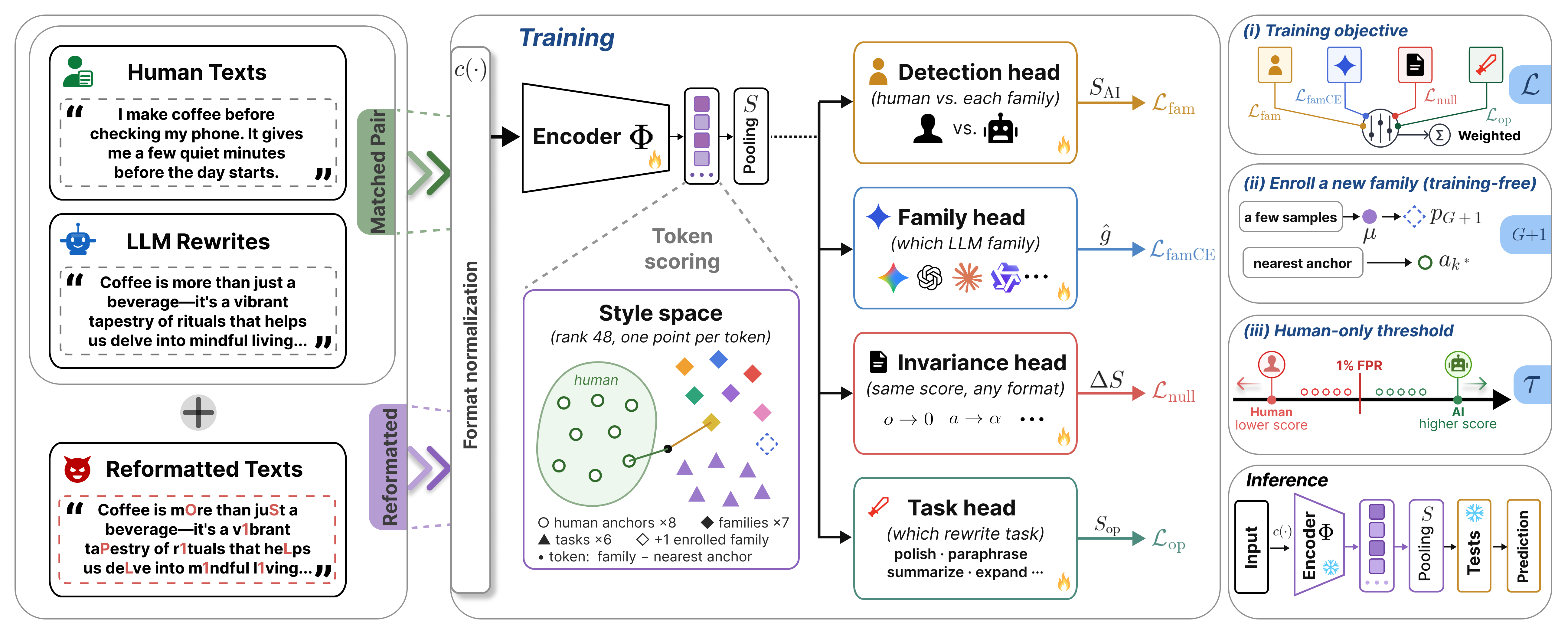}
\caption{\textbf{Overview of MELD.} \emph{Left:} each human text passage
is paired with an AI rewrite of the same content and with reformatted
copies, and every input undergoes format normalization first.
\emph{Middle:} a shared encoder maps tokens into a style space with
eight human anchors and one center per generator family; the most
AI-like tokens are pooled into the document score, and four heads are
trained for family-balanced detection, family attribution, format
invariance, and rewrite-task attribution. \emph{Right:} the four losses
form one objective, a new family is enrolled from a few samples without
retraining, and the threshold is set from human text alone; at inference,
one pass gives the detection score and the family, task, and span-level
predictions.}
\label{fig:overview}
\end{figure}
\section{Introduction}
\label{sec:intro}
As large language models become common in everyday writing, reliable AI-text detection is increasingly important for academic integrity, content moderation, and provenance. In real-world use, overall accuracy alone is not enough. False positives can carry
serious consequences for human authors, including accusations of academic
misconduct and unfair penalties for non-native English writers
\citep{liang2023gpt}. Simple paraphrasing and rewriting strategies have
also been shown to evade or destabilize existing detectors
\citep{krishna2023paraphrasing,hu2023radar,wu2024detectrl}.

Current methods fall into three categories
(Section~\ref{sec:related}): 1) training-free detectors use token-rank,
likelihood-curvature, or cross-perplexity signals from reference language
models
\citep{gehrmann2019gltr,mitchell2023detectgpt,bao2023fast,hans2024spotting};
2) supervised encoder detectors learn binary classifiers from labeled
examples through a single binary objective
\citep{solaiman2019release,guo2023close,drayson2025modernbertdetect}; and
3) a newer line incorporates fine-grained authorship or generator
structure through contrastive learning, auxiliary supervision,
representation analysis, and out-of-distribution detection
\citep{guo2024detective,ta2026faid,chen2025repreguard,zeng2025human}.
These advances improve benchmark performance, but high overall AUROC
can hide poor performance on rare, difficult, or unseen
generators. A pooled AI-versus-human objective treats all generators as one “AI” class, providing no explicit supervision for generator- or rewrite-specific differences. As a result, the model may learn features sufficient for aggregate detection without capturing the finer structure needed for attribution and adaptation to new families. Strong performance on clean text also does
not ensure reliable detection at low false-positive rates after
rewriting or reformatting \citep{dugan2024raid}. Three challenges
therefore remain: combining detection with family and task attribution,
maintaining robust detection at low false-positive rates under attack,
and generalizing to unseen generators while enabling efficient enrollment of new families.

We propose MELD (Multi-Task Equilibrated Learning Detector), a detector
that trains one class-balanced test per generator family instead of a
single pooled AI-versus-human classifier. Each test distinguishes that
family's text from human writing, and the family losses receive equal
weight, making each family's detection task an explicit part of
training. All tests share one model of human writing: a fine-tuned
encoder maps each token into a small style space with several human
anchors and one learned center per family. Each token receives a score
based on its distances to a family center and the nearest human anchor,
and a document score averages its most AI-like tokens
(Figure~\ref{fig:overview}). The same head is trained for generator-family
and rewrite-task attribution and also provides predictions of AI-written
spans. To support robust detection at low FPR, we train on AI rewrites,
encourage consistent scores across formatting variants, and use format
normalization to make scores invariant to different formatting
changes. Detection thresholds are calibrated from representative human
text alone. To encourage generalization, the training data include
$1.25$M AI rewrites of pre-2021 arXiv passages that match human passages
in topic, field, length, and mathematical content, helping the detector
learn how a text is written rather than what it is about. The
family-center design also allows a new generator family to be enrolled
from a few examples without retraining or changing the existing family
tests. Our main contributions are as follows:
\begin{list}{\textbullet}{
\setlength{\leftmargin}{1.1em}
\setlength{\itemsep}{1pt}
\setlength{\topsep}{2pt}
\setlength{\parsep}{0pt}
\setlength{\partopsep}{0pt}
}
\item \textbf{Multi-task family-aware detection.}
MELD uses class-balanced, family-specific detection tasks over a shared model of human writing, while jointly supporting family attribution, rewrite-task prediction, and AI-span localization. Its scoring design also supports format invariance, human-only threshold calibration, and few-shot enrollment of new families.

\item \textbf{MELD-eval.}
We introduce a held-out evaluation pool built from four current-generation chat models, paired human text, and RAID-style domains and attacks to test zero-shot transfer to unseen generators.

\item \textbf{Strong and robust performance.}
MELD ranks first among open-source systems on the RAID leaderboard, achieves 99.7\% TPR at 1\% FPR on MELD-eval without further fine-tuning, and reaches 92\% top-1 accuracy for seven-way family attribution. We further apply MELD to 5.9M scientific texts from 2016--2026.
\end{list}

\begin{table*}[tb]
\centering
\small
\setlength{\tabcolsep}{4pt}
\caption{\textbf{RAID public leaderboard} (\url{https://raid-bench.xyz}, accessed 2026-09-23). AUROC and TPR at $5\%$/$1\%$ FPR ($\times100$) on the official RAID test set. ``All settings'' includes RAID's attack suite; ``No attack'' is the clean subset. Commercial rows are public product submissions; open-source rows are leaderboard submissions with a paper and public model or code. ``---'': attacked scores unavailable. \textbf{Best}/\underline{second-best} per column.}
\label{tab:raid}
\input{figures/main_table}
\end{table*}

\section{Related work}
\label{sec:related}

\paragraph{Training-free detectors.}
Training-free methods usually score text under one or more reference language models (LMs) and use token statistics, likelihood geometry, or cross-model discrepancies as evidence of generation. GLTR \citep{gehrmann2019gltr} uses token-rank statistics. DetectGPT \citep{mitchell2023detectgpt} and Fast-DetectGPT \citep{bao2023fast} rely on likelihood curvature. Binoculars \citep{hans2024spotting} compares cross-perplexities from two LMs. These detectors are easy to deploy because they do not require detector-specific training, but their behavior is tied to the coverage and calibration of the reference models, making them sensitive to paraphrase and surface perturbations \citep{krishna2023paraphrasing,dugan2024raid}.

\paragraph{Supervised detectors.}
Supervised methods train discriminative models from labeled human and AI text. Early studies finetuned RoBERTa-style encoders \citep{solaiman2019release,guo2023close}. Subsequent work improved this recipe with structured features \citep{verma2024ghostbuster}, adversarial paraphrasing \citep{hu2023radar}, stronger encoder backbones \citep{warner2025smarter,drayson2025modernbertdetect}, representation-based detection \citep{chen2025repreguard,detectanyllm2025}, and one-class objectives \citep{zeng2025human}. While these methods can perform well on in-distribution benchmarks, they are typically trained with a single binary head, so their pooled scores can mask poor performance on individual generator families.

\paragraph{Generator structure in detectors.}
Recent detectors use generator-aware contrastive learning \citep{guo2024detective}, auxiliary tasks \citep{ta2026faid}, few-shot style representations \citep{soto2024fewshot}, family-aware mixtures of experts \citep{phantomhunter2025}, semantic subspaces \citep{prossd2026}, or prototypes for both classes under a pooled decision \citep{xu2026samp}. MELD instead lets family identity define the tests, with a multi-anchor human model shared across them, and follows established practice for its detection-plus-attribution predictions \citep{phantomhunter2025,ta2026faid}. Group-DRO reweights weak groups \citep{sagawa2020distributionally}, as do fairness objectives for visual deepfake detection \citep{ju2024improving}; MELD keeps the group structure with uniform weights (Section~\ref{sec:objective}).

\paragraph{Invariance and behavioral tests.}
Homoglyph attacks can change text-detector verdicts without changing appearance \citep{wolff2020attacking,creo2025silverspeak}. Learned normalization for equivariance \citep{kaba2023canonicalization,mondal2023equivariant}, scale-space equivariance \citep{worrall2019deep}, transformation-invariant pooling \citep{laptev2016ti}, and robust encodings for typos \citep{jones2020robust} assume group-structured or enumerated transformation sets; MELD's format normalization absorbs the modeled formatting changes (Section~\ref{sec:deploy}). Counterfactual invariance formalizes stability under changes to spurious attributes \citep{veitch2021counterfactual,fawkes2023results}, and human originals paired with AI rewrites appear in ImBD \citep{chen2025imitate}, rewrite-distance detection \citep{hao2025learning}, and matched-rewrite benchmarks \citep{perrone2026arb,bsharat2026opai}. Behavioral tests separate changes that should and should not affect predictions \citep{ribeiro2020beyond,goodfellow2009measuring,lovering2021predicting,tiwari2026reward}; our two contrasts apply both kinds of change to the same document (Section~\ref{sec:sar}).

\section{MELD - Model, Method and Deployment}
\label{sec:method}

\subsection{Family-specific detection with a shared human model}
\label{sec:setup}
Let $x$ denote a document with label $y\in\{0,1\}$, where $y=1$ indicates AI-generated text. Each AI document has a generator-family label $g(x)\in\{1,\ldots,G\}$. A family groups related generators, such as GPT or Llama; we map the $78$ training generator identifiers to $G{=}7$ families. MELD learns one binary test per family, distinguishing that family's text from human writing. All tests share one human model. At inference, their outputs form one detection score. We track each test during training and evaluation and report the worst-performing family alongside aggregate results.

\subsection{Style representations and scoring}
\label{sec:geometry}
We use Ettin-1B \citep{weller2025seq}, a ModernBERT-family encoder \citep{warner2025smarter} with hidden size $1792$ and an $8192$-token context. A lightweight head (under $0.01\%$ of the encoder's parameters) projects tokens into a $48$-dimensional \emph{style space}. We jointly train the projection and encoder to distinguish human and AI writing (Section~\ref{sec:objective}).

Human writing has $M{=}8$ learned anchors in this space, and each generator family has one center. Affinities are negative squared distances scaled by a learned temperature, with a learned bias per family. A token's family score is its affinity to that family's center minus its affinity to the nearest human anchor, allowing different human styles to match different anchors. We clip these scores to $[-\kappa,\kappa]$, with $\kappa{=}30$, to bound individual token contributions \citep{huber1965robust}. A log-sum-exp over the $G$ family scores gives a combined AI score for each token.

Document scores average the highest $q=\lceil\rho n\rceil$ scores among $n$ valid tokens, with $\rho{=}0.25$. Pooling family-specific scores gives $\Sfam[g]$; pooling combined scores gives $\Sai$. This multiple-instance rule \citep{durand2016weldon} reduces the dilution of localized AI content by surrounding human text and requires only document-level supervision. Affinities to six learned task centers (generate, polish, paraphrase, humanize, summarize, expand) use the same pooling to give $S_{\mathrm{op}}$. An energy score $\Senergy$ over all valid tokens \citep{liu2020energy} is low when representations are far from the learned anchors and centers. Appendix~\ref{app:geometry} gives the complete definitions.

\subsection{Training objective}
\label{sec:objective}
We train each pooled family score $\Sfam[g]$ with binary cross-entropy, using that family's AI documents as positives and all human documents in the batch as negatives. We average the two class means with equal weight, then average the family losses: $\mathcal{L}_{\mathrm{fam}}=\tfrac1G\sum_{g=1}^{G}\mathcal{L}_g$. Softmax cross-entropy trains family attribution from family scores on AI examples ($\mathcal{L}_{\mathrm{famCE}}$), while six-way cross-entropy trains task scores ($\mathcal{L}_{\mathrm{op}}$). On $15\%$ of training examples, a consistency loss $\mathcal{L}_{\mathrm{null}}$ applies a Huber penalty to the score difference between two randomly formatted versions of the same text, encouraging stable raw scores before format normalization. The full objective is
\begin{equation}
\mathcal{L}=\mathcal{L}_{\mathrm{fam}}+0.15\,\mathcal{L}_{\mathrm{op}}+0.15\,\mathcal{L}_{\mathrm{famCE}}+0.10\,\mathcal{L}_{\mathrm{null}}.
\label{eq:objective}
\end{equation}
Human passages and paired rewrites enter as individual examples, without a pairwise loss (Section~\ref{sec:data}). Matching topic, field, section role, length band, and mathematical content across labels encourages learning differences in the writing process. Appendix~\ref{app:stat_sig} ablates each loss.

\subsection{Deployment properties}
\label{sec:deploy}
\paragraph{Format invariance.}
\looseness=-1
Look-alikes, invisible characters, whitespace, and markdown can alter predictions without substantially changing displayed text \citep{wolff2020attacking,creo2025silverspeak}. MELD normalizes look-alikes, removes invisible characters and markup, and collapses whitespace. For modeled operations, $R_k=\operatorname{format}_k\circ\operatorname{normalize}$ satisfies $R_k\circ R_j=R_k$ for all $j,k$. Reformatted versions share one normalized form and score. Unlike group-based approaches \citep{kaba2023canonicalization,mondal2023equivariant}, no group is required. We observed no violations in $484{,}396$ compositions; format normalization completely absorbs RAID's homoglyph and zero-width attacks. Robustness to wording changes such as paraphrasing is learned from training data.

\paragraph{Bounded token influence.}
For a fixed-length token-score vector, changing at most $m$ entries changes the pooled score by at most $2\kappa m/\lceil\rho n\rceil$ (Proposition~\ref{prop:influence}, Appendix~\ref{app:proofs}).

\paragraph{Human-only threshold calibration.}
Target-FPR thresholds are human-score quantiles from a held-in calibration set, stored with the checkpoint. We recalibrate on representative human text for new deployment populations (Section~\ref{sec:study}); no AI examples are needed.

\paragraph{New-family enrollment.}
We enroll a family without retraining by initializing its center from about $64$ documents' mean style representation and adjusting it away from the nearest human anchor (Appendix~\ref{app:enroll}). The encoder, human anchors, and existing tests remain unchanged. If the new family's token scores stay sufficiently below the existing combined scores, the increase in combined score is exponentially small (Proposition~\ref{prop:enroll}, Appendix~\ref{app:proofs}).

\section{Data and training}
\label{sec:data}

\paragraph{Training mixture.}
One of our main efforts is to curate a comprehensive training corpus and MELD-eval, a held-out transfer benchmark (Appendix~\ref{app:meld_eval}). The training mixture combines our rewrites with public datasets: $10.30$M examples, one schema, and seven generator families (Appendix~\ref{app:mixture}, Table~\ref{tab:data}). Nine human registers, including scientific prose, peer reviews, and news, define calibration groups and evaluation subsets. Perturbations keep source family labels; sampling rebalances family counts. Validation has $50{,}128$ examples across all families. Deduplication against a fixed evaluation registry removes $104{,}660$ near-duplicates; arXiv identifiers exclude training papers from the case study (Section~\ref{sec:study}).

\paragraph{Paired scientific rewrites.}
We construct the scientific rewrite corpus from a fixed snapshot of $3.09$M papers' \LaTeX{} sources. Every recorded paper version must predate 1~January 2021. Four open-weight models use nine operations, mapped to six training tasks (Appendix~\ref{app:rewrite}). The $3.07$M examples comprise $1{,}476{,}558$ source--rewrite pairs and $115{,}155$ unpaired generation examples. Paired passages share topic, field, and section role, meet length constraints, and preserve all equations and citation keys.

\paragraph{Training procedure.}
Length-bucketed batches with a fixed token budget support up to $8192$ tokens per document and are $50\%$ human. Random cropping, line-wrap randomization, markdown removal/insertion, and light post-editing apply to both labels to reduce formatting and editing cues. Encoder and head are jointly fine-tuned for one $7.06$M-example sampling epoch on eight L40 GPUs (Appendix~\ref{app:training}).

\section{Results and Discussion}
\label{sec:results}

\subsection{Validation and unseen-family tests}
\label{sec:val}
On the $50{,}128$-example validation split, MELD achieves $99.78$ AUROC and $97.5\%$ TPR at $1\%$ FPR. Worst-family AUROC is $99.68$, only $0.10$ points below the overall value. The stored nominal-$1\%$ threshold yields $1.6\%$ FPR on validation ($5.5\%$ for texts of at most $192$ words), $0.7\%$ on paired arXiv, $0.3\%$ on the DeepSeek evaluation set, and $0.0$--$1.9\%$ across registers. Enumeration-heavy human text, including author lists and bibliographies, stays below $0.1\%$ FPR.

DeepSeek-V4-Flash and -Pro, whose family is absent from training, both reach $99.9$ AUROC, with $98.0\%$ and $98.3\%$ TPR at $1\%$ FPR. The in-mixture gpt-oss-20b control reaches $99.82$ AUROC on the same source passages and rewrite pipeline (Appendix~\ref{app:rewrite}). Holding these fixed shows strong transfer to the held-out generators under this protocol. On Beemo, whose AI text is expert-edited, MELD reaches $81.77$ AUROC versus $51$--$69$ for three public baselines (Appendix~\ref{app:stat_sig}).

Family attribution reaches $91.9\%$ top-1 across seven families on $8{,}400$ validation examples sampled by writing category, with at least $84.3\%$ in every register, indicating consistent performance across the evaluated registers.

\subsection{Public benchmarks}
\label{sec:bench}
Baselines use official code or public checkpoints where available. AUROC is on a $0$--$100$ scale; TPR thresholds are calibrated on each pool's human scores, rather than shared across pools. Tables~\ref{tab:bench} and~\ref{tab:meld_eval} report $95\%$ percentile bootstrap CI half-widths ($B{=}1{,}000$; Appendix~\ref{app:protocol}).

\paragraph{RAID leaderboard.}
On RAID \citep{dugan2024raid}, MELD has the strongest open-source results and is competitive with commercial detectors (Table~\ref{tab:raid}). Across all settings, including attacks, its mean AUROC, TPR@$5\%$FPR, and TPR@$1\%$FPR exceeds the next-best open-source mean by $6.3$ points and the highest commercial mean by $0.4$.

\begin{table*}[tb]
\centering
\small
\setlength{\tabcolsep}{3pt}
\caption{\textbf{Held-out pool AUROC} ($\times100$). Baselines use public checkpoints or official code on the same pools, all excluded from MELD training (Section~\ref{sec:data}). Entries give the estimate $\pm$ half-width of a $95\%$ bootstrap CI. \textbf{Best}/\underline{second-best} per column; RoBERTa-Large is OpenAI's detector.}
\label{tab:bench}
\input{figures/tab_benchmarks}
\end{table*}

Table~\ref{tab:bench} compares reference-LM, supervised, and representation-based detectors on HC3 \citep{guo2023close}, MAGE \citep{li2024mage}, M4GT \citep{wang2024m4gt}, Ghostbuster \citep{verma2024ghostbuster}, and DetectRL \citep{wu2024detectrl}, all excluded from training (Section~\ref{sec:data}). MELD has the highest AUROC point estimate on four of five pools. On HC3, which is closer to RoBERTa-ChatGPT's training distribution, that detector reaches $100.0$ versus MELD's $99.90$. On M4GT, MELD's $77.5\pm1.3$ and the next-best $76.9\pm1.8$ have overlapping intervals.

\paragraph{MELD-eval.}
\looseness=-1
GPT-5.4-Mini, Gemini-3-Flash, Claude-Haiku-4.5, and Qwen-3.6-Plus \citep{openai2026gpt54mini,google2026gemini3flash,anthropic2025claudehaiku45,alibaba2026qwen36plus} generate text under one no-markdown instruction across eight RAID English domains, followed by RAID-style attacks. The constructed pool has $7{,}862$ paired human texts, $31{,}448$ clean AI examples, and $188{,}688$ attacked examples (Appendix~\ref{app:meld_eval}).

\begin{table}[tb]
\centering
\small
\caption{\textbf{MELD-eval TPR@$1\%$FPR} ($\times100$), per generator and overall, against paired human texts. All detectors are zero-shot for these generators. Entries give the estimate $\pm$ half-width of a $95\%$ bootstrap CI. \textbf{Best}/\underline{second-best} per column.}
\label{tab:meld_eval}
\setlength{\tabcolsep}{4pt}
\input{figures/tab_meld_eval}
\end{table}

Without further fine-tuning, MELD reaches at least $99.3\%$ TPR at $1\%$ FPR per generator and $99.7\%$ overall (Table~\ref{tab:meld_eval}). ModernBERT-Detect is the strongest baseline at $95.5\%$ overall under the same calibration protocol; most others are substantially lower. MELD is also more consistent across generators. Appendix~\ref{app:cases} gives HC3, DetectRL, and MELD-eval examples.

\subsection{Process and surface sensitivity}
\label{sec:sar}
Evasion tests measure AI text escaping detection \citep{wolff2020attacking,creo2025silverspeak,dugan2024raid}; they do not directly test human false positives under reformatting or compare the same edits across both classes. On held-out documents, the \emph{process contrast} $\Delta_p=S(x^{\mathrm{ai}})-S(x^{\mathrm{h}})$ compares content-matched AI rewrites and human sources. The \emph{surface contrast} $\Delta_s=S(R_kx)-S(x)$ compares eight formats, including plain text, headings, bold, lists, fenced code, tables, and chat layouts, with content and authorship fixed. Their dimensionless ratio, $\mathrm{SAR}=\mathrm{median}|\Delta_p|/\mathrm{median}|\Delta_s|$, applies to any scalar detector, including black boxes. Below one, median absolute score changes are larger for formatting than rewriting.

\begin{table}[tb]
\centering
\small
\setlength{\tabcolsep}{4pt}
\caption{\textbf{Two contrasts: 400 held-out arXiv pairs $\boldsymbol{\times}$ 8 formats.} $\pm$: $95\%$ bootstrap CI half-width ($B{=}1{,}000$). Worst FPR (\%) uses each detector's $1\%$-FPR threshold before/after format normalization. RoBERTa-Large is OpenAI's detector. \textbf{Best}/\underline{second-best}: AUROC, SAR, FPR.}
\label{tab:sar}
\begin{tabular*}{\textwidth}{@{\extracolsep{\fill}}lrrrrcc@{}}
\toprule
\textbf{Detector} & \textbf{Pair AUROC}$\uparrow$ & \textbf{med.\ $|\Delta_p|$} & \textbf{med.\ $|\Delta_s|$} & \textbf{SAR}$\uparrow$ & \textbf{Raw FPR}$\downarrow$ & \textbf{Norm. FPR}$\downarrow$ \\
\midrule
RoBERTa-ChatGPT              & 63.85 & \tableci{0.89}{0.19} & \tableci{0.35}{0.03} & \underline{\tableci{2.5}{0.6}} & 4.3 & \textbf{0.8} \\
RoBERTa-Large                & 63.56 & \tableci{0.97}{0.12} & \tableci{0.54}{0.02} & \tableci{1.8}{0.2} & \textbf{0.5} & \textbf{0.8} \\
ModernBERT-Detect            & \underline{81.67} & \tableci{1.64}{0.20} & \tableci{1.87}{0.08} & \tableci{0.9}{0.1} & 7.7 & \textbf{0.8} \\
\midrule
\textbf{MELD} (ours)         & \textbf{99.97} & \tableci{10.27}{0.16} & \tableci{0.32}{0.02} & \textbf{\tableci{31.7}{2.3}} & \underline{1.8} & \underline{1.0} \\
\bottomrule
\end{tabular*}
\end{table}

\begin{figure}[tb]
\centering
% Clip only blank top/bottom borders when enlarging the result figures.
\includegraphics[width=0.8\textwidth,trim=0 22.66bp 0 11.66bp,clip]{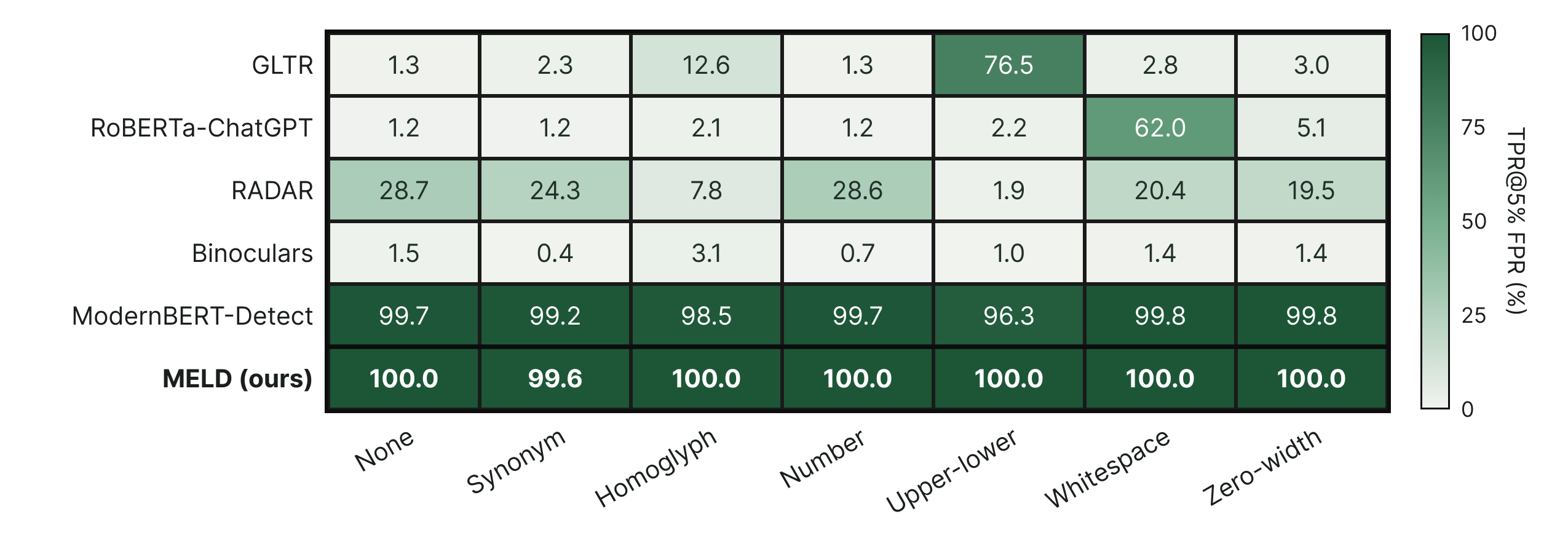}
\caption{\textbf{Per-attack detection on MELD-eval.} TPR@$5\%$FPR for the clean setting and each of the six attack types, for MELD and five baselines; every detector's threshold is set once on the paired human rows and held fixed across attacks.}
\label{fig:per_attack}
\end{figure}

With format normalization disabled, Table~\ref{tab:sar} compares $400$ held-out arXiv pairs across eight formats. MELD's SAR is $31.7$, and every AI rewrite outscores its source. ModernBERT-Detect's SAR is $0.9$, with worst-format human FPR of $7.7\%$: detection accuracy does not ensure format stability. Format normalization yields worst-format FPRs of $0.8$--$1.0\%$ across all four detectors. MELD's deployed pipeline gives $\Delta_s=0$, so SAR is undefined (Appendix~\ref{app:protocol}). Figure~\ref{fig:per_attack} complements these contrasts with per-attack detection under a fixed threshold.

\subsection{Representation analysis}
\label{sec:repr}
\begin{figure}[tb]
\centering
\includegraphics[width=1.0\textwidth,trim=0 4.58bp 0 5.56bp,clip]{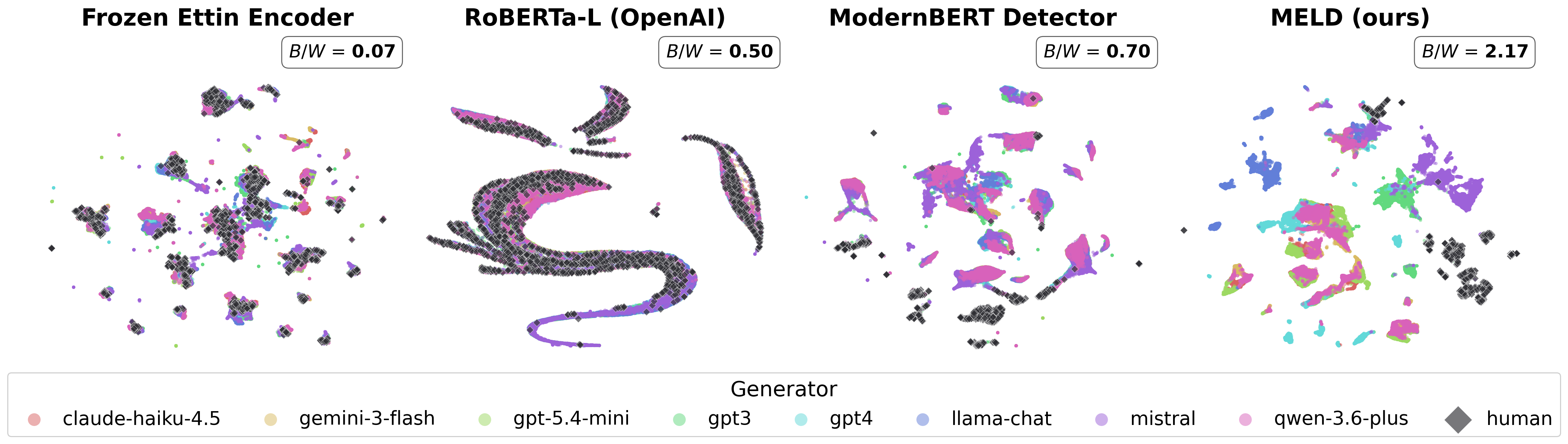}\\[0.3em]
\includegraphics[width=1.0\textwidth,trim=0 17.01bp 0 5.56bp,clip]{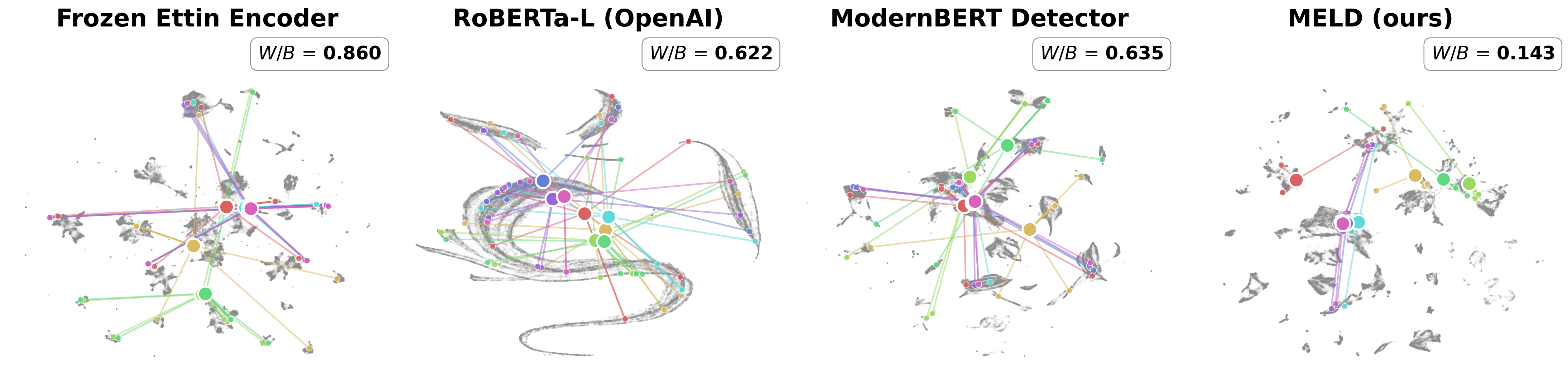}
\caption{\textbf{Backbone geometry.} UMAP of $144{,}000$ embeddings per panel. \emph{Top:} generator colors, with human text in black; between-/within-generator distance $B/W$ is $2.17$ for MELD and at most $0.70$ for baselines. \emph{Bottom:} Original text joined to the attacked variants; within-/between-source distance $W/B$ is $0.14$ for MELD versus $0.62$ for the nearest baseline.}
\label{fig:embedviz_combined}
\end{figure}

\begin{figure}[tb]
\centering
\includegraphics[width=1.0\textwidth,trim=0 4.58bp 0 5.56bp,clip]{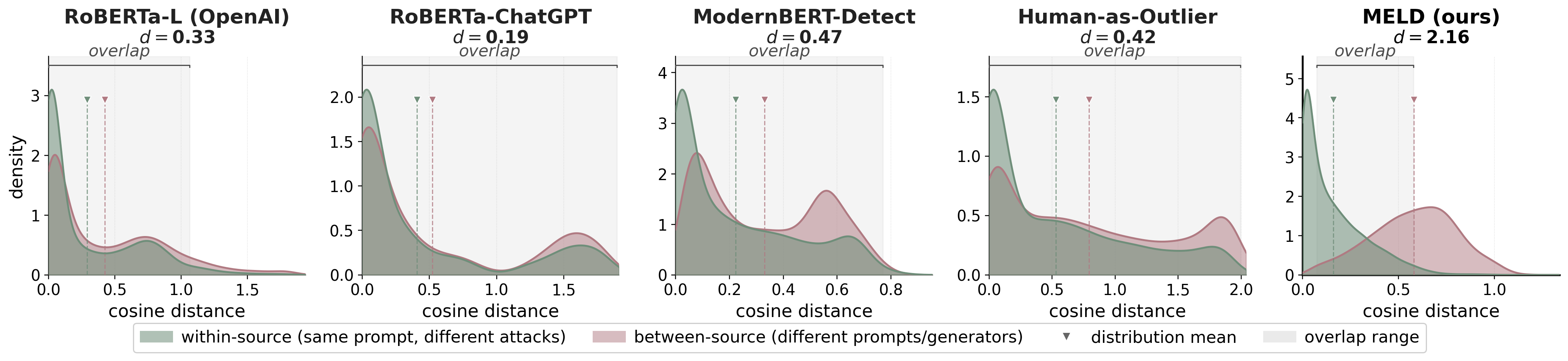}
\caption{\textbf{Distance-space geometry.} \textcolor[RGB]{111,142,122}{\textbf{Within-source}} and \textcolor[RGB]{176,122,130}{\textbf{between-source}} cosine-distance distributions; $d$ is their standardized gap (MELD $2.16$, next ModernBERT-Detect $0.47$).}
\label{fig:dist_hist}
\end{figure}

We analyze $144{,}000$ embeddings per detector (Figures~\ref{fig:embedviz_combined} and~\ref{fig:dist_hist}). MELD's between-/within-generator distance ratio is $2.2$ versus the best baseline's $0.7$; under attack, its within-/between-source ratio is $0.14$ versus the nearest baseline's $0.62$. The standardized within-/between-source distance gap is $d{=}2.16$ versus ModernBERT-Detect's $0.47$. These results are consistent with greater generator separation and stability under the evaluated attacks.

\subsection{Limitations}
\label{sec:limitations}
Evaluation covers English documents that are fully AI-generated or AI-edited; other languages and writing settings require corresponding data. TPR-at-FPR results use each pool's human-score thresholds, so a stored threshold need not retain its FPR across domains, registers, or lengths. Deployment requires calibration on representative human text (Section~\ref{sec:study}). Format normalization guarantees exact invariance to the modeled formatting transformations. Robustness to wording changes, including paraphrasing and humanization, is learned from training data and evaluated empirically on rewrite and attack benchmarks.

\section{Case study: scientific writing and peer review}
\label{sec:study}

We use MELD's scores and task predictions to track signs of AI assistance across years, fields, paper sections, and peer reviews, examining whether flagged text resembles AI editing or generation from scratch. We analyze $5.9$ million texts: $5{,}682{,}241$ section units from $1{,}344{,}610$ arXiv papers (2016--2026), $92{,}174$ accepted abstracts from nine conferences, and $143{,}685$ review summaries. We keep the MELD checkpoint with nine operation labels fixed, exclude training papers by identifier, and match review fields across years (Appendix~\ref{app:study_details}). Separate 2016--2022 baselines for each corpus, section type, venue, and field define score percentiles; texts above the $99$th percentile are \emph{flagged}. $P(\mathrm{AI})=\sigma(\Sai)$ rescales the score; it is not the probability that the text was written by AI.

\begin{figure}[t]
\centering
\captionsetup{skip=3pt}
\includegraphics[width=0.80\textwidth]{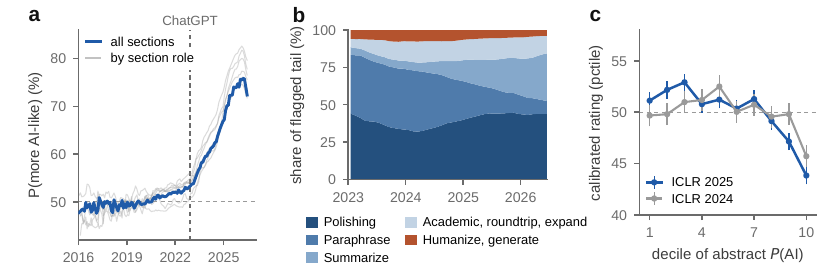}
\caption{\textbf{Score trends, predicted writing tasks, and review ratings.}
\emph{(a)}~Monthly probability that an arXiv section receives a higher
MELD score than a baseline section of the same type.
\emph{(b)}~Predicted writing tasks among flagged arXiv sections, shown
as percentages over time.
\emph{(c)}~Average ICLR review ratings for papers divided into ten groups
by their abstracts' $P(\text{AI})$ scores, from lowest to highest.
Ratings are converted to percentiles within each year, and error bars
show one standard error.}
\label{fig:study_transition}
\end{figure}

\paragraph{Changes over time.}
Scores stay relatively stable through 2022 and rise from 2023 (Table~\ref{tab:study_year}; Figure~\ref{fig:study_transition}a). The flagged arXiv share grows from $4.1\%$ in 2023 to $25.4\%$ in January--July 2026, about $24$ times baseline, with a median shift of $0.72$ baseline standard deviations (SD). Equal yearly samples and first submissions retain this pattern (Appendix~\ref{app:study_details}), suggesting that sample growth or revisions alone do not explain it. For most flagged texts, MELD predicts polishing or paraphrasing; fewer than $2\%$ are assigned generation from scratch (Figure~\ref{fig:study_transition}b). In some AI-related fields, scores plateau or fall in January--July 2026 as energy scores change, possibly reflecting changing generator styles rather than less AI use.

\paragraph{Sections and authors.}
By 2026, conclusions, discussions, and related work shift $1.6$--$3.4$ baseline SD, versus less than one for methods, results, and abstracts. Timing is similar across fields and nine venues. Among $23{,}404$ first authors publishing before 2023 and from 2023 onward, $77\%$ have higher later score percentiles (Appendix~\ref{app:study_details}).

\paragraph{Peer review and outcomes.}
In 2025, median review-summary scores exceed those of $87$--$88\%$ of matched baseline reviews; ICLR summaries are flagged $1.7$ times as often as arXiv units using separate baselines. Flagged and unflagged reviews give similar rating percentiles within venue and year. Papers whose abstracts have the highest detector scores receive lower ratings (Figure~\ref{fig:study_transition}c); acceptance is roughly flat across the first nine abstract-score deciles and lower in the highest (Appendix~\ref{app:study_details}).

\section{Conclusion}
\label{sec:conclusion}
We introduced MELD, a multi-task model that detects AI-generated text
and predicts its generator family and rewrite task. It provides
document- and token-level predictions in a single forward pass,
remains invariant to the formatting changes, and allows
new generator families to be enrolled without large retraining. We also introduced MELD-eval, a held-out benchmark covering four frontier chat models and different attacks. MELD ranks first among open-source submissions on the RAID leaderboard and remains competitive with commercial systems. It achieves $99.7\%$ TPR at $1\%$ FPR on MELD-eval without further fine-tuning and $98\%$ TPR at the same FPR on a generator whose family is absent from training. Finally, a case study of $5.9$ million scientific texts from 2016--2026 finds that detector scores remain relatively stable through 2022 and increase from 2023 onward.

\clearpage
\bibliographystyle{iclr2027_conference}
\bibliography{refs}

\clearpage
\appendix

\section{Scoring geometry and objective}
\label{app:geometry}

Let $x$ be a document. MELD combines $G$ family-specific detection
tests that share a human null hypothesis. For AI-generated text,
$g(x)\in\{1,\ldots,G\}$ denotes its generator family. The tests compare
\[
    \mathcal{H}_0: x \text{ is human-written}
    \qquad\text{against}\qquad
    \mathcal{H}_1^{(g)}: x \text{ is generated by family } g,
    \quad g=1,\ldots,G.
\]

Let $\Phi:\mathcal{X}\to\mathbb{R}^{L\times H}$ be a bidirectional
encoder that produces token representations $h_\ell\in\mathbb{R}^H$.
We use Ettin-1B \citep{weller2025seq}, a ModernBERT-family encoder
\citep{warner2025smarter} with hidden size $H=1792$ and a context
length of $8192$ tokens. A learned projection $B$ maps each token
representation into a style space of dimension $r=48$:
\[
    u_\ell
    =\operatorname{LayerNorm}(B^\top h_\ell)
    \in\mathbb{R}^r.
\]
The head contains $87{,}134$ parameters, less than $0.01\%$ of the
backbone's parameter count.

Human writing is represented by $M=8$ anchors $a_k$, and each
generator family has a center $p_g$. Their affinities to a style
representation $u$ are
\[
\begin{aligned}
    \varphi^{\mathrm{h}}_k(u)
    &= -\tau\lVert u-a_k\rVert^2, \\
    \varphi^{\mathrm{g}}_g(u)
    &= -\tau\lVert u-p_g\rVert^2+b_g,
\end{aligned}
\]
where $\tau$ is a learned temperature and $b_g$ is a learned
family-specific bias. The human affinity is the affinity to the
nearest human anchor:
\[
    \operatorname{hum}(u)
    =\max_{k\le M}\varphi^{\mathrm{h}}_k(u).
\]
We define the family-specific and combined token scores as
\begin{equation}
\begin{aligned}
    s_{\ell g}
    &=\operatorname{clip}\!\left(
        \varphi^{\mathrm{g}}_g(u_\ell)
        -\operatorname{hum}(u_\ell),
        -\kappa,\kappa
      \right), \\
    c_\ell
    &=\tau_{\mathrm{agg}}\,
      \LSE_{g\le G}\!\left(s_{\ell g}/\tau_{\mathrm{agg}}\right),
    \qquad
    \LSE_g z_g=\log\sum_g e^{z_g}.
\end{aligned}
\label{eq:tokenlr}
\end{equation}
We use $\kappa=30$ and $\tau_{\mathrm{agg}}=1$. The margin
$s_{\ell g}$ compares family $g$ with the shared human model,
while $c_\ell$ combines evidence across families. Clipping bounds
each family-specific token margin \citep{huber1965robust}.
These affinity contrasts are a discriminatively trained analogue
of the likelihood-ratio scores used with universal background
models in speaker verification \citep{reynolds2000speaker}.

Document scores use top-quantile pooling
\citep{durand2016weldon}. For a vector $v$ of $n$ valid token
scores, let $I_q(v)$ index its $q=\lceil\rho n\rceil$ largest
entries, where $\rho=0.25$. Define
\[
    \operatorname{TQM}_\rho(v)
    =\frac{1}{q}\sum_{\ell\in I_q(v)}v_\ell.
\]
The family score $\Sfam[g]$ applies this operation to the
family-$g$ token margins, and the detection score $\Sai$ applies
it to the combined token scores $c_\ell$.

The same style space supports two additional outputs. Six task
centers represent generate, polish, paraphrase, humanize,
summarize, and expand. Their token affinities use the same
pooling rule to produce $S_{\mathrm{op}}\in\mathbb{R}^6$.
The energy score $\Senergy$ \citep{liu2020energy} averages
log-sum-exp affinity over the $n$ valid tokens:
\[
    \Senergy=\frac{1}{n}\sum_{\ell=1}^{n}
    \log\!\left(
        \sum_{g=1}^{G}e^{\varphi^{\mathrm{g}}_g(u_\ell)}
        +\sum_{k=1}^{M}e^{\varphi^{\mathrm{h}}_k(u_\ell)}
    \right).
\]
Lower values indicate representations far from the learned
human anchors and family centers.

For document $i$, the pooled family score $\Sfam[i,g]$ serves
as the binary-classification logit. The loss for family $g$ is
\begin{equation}
\begin{aligned}
    \mathcal{L}_g
    =\frac{1}{2}\Big[&
        \operatorname{mean}_{i\in P_g}
        \operatorname{BCE}\!\left(\Sfam[i,g],1-\varepsilon\right)
        \\
        &+\operatorname{mean}_{i\in N}
        \operatorname{BCE}\!\left(\Sfam[i,g],0\right)
      \Big],
\end{aligned}
\label{eq:lg}
\end{equation}
where $P_g$ contains the batch's AI examples from family $g$
and $N$ contains its human examples. We use
$\varepsilon=0.02$ to smooth positive targets only. If a batch
contains no examples from family $g$, the positive term is
omitted. The family losses receive equal weight:
\[
    \mathcal{L}_{\mathrm{fam}}
    =\frac{1}{G}\sum_{g=1}^{G}\mathcal{L}_g.
\]
The term \emph{equilibrated} refers to the class-balanced family
losses. The auxiliary losses and their weights are given in
Section~\ref{sec:objective}.

\paragraph{Span and task predictions.}
When a token has zero margin for every family, its combined
score is $\tau_{\mathrm{agg}}\log G$. A two-state Viterbi decoder
uses this neutral value to segment a mixed document into human
and AI spans. Family and task scores provide the corresponding
span attributions. No span-level labels are used during training.
Appendix~\ref{app:stat_sig} evaluates mixed-document detection
using pooled document scores.
Writing $e_\ell=c_\ell-\tau_{\mathrm{agg}}\log G$, the decoder
uses emissions $-e_\ell$ for the human state and $+e_\ell$ for
the AI state, with a cost of $4$ for each state change.

\paragraph{Human-only thresholds.}
For a target false-positive rate, we select the corresponding
quantile of human detection scores. The calibration pool contains
human texts from nine writing categories (registers), including
scientific papers, peer reviews, and news. We store the resulting
thresholds in the checkpoint. No AI-generated examples are required.
When the deployment population differs from the calibration
pool, thresholds should be recomputed from representative local
human text, as in Section~\ref{sec:study}.

The quantity $P(\text{AI})=\sigma(\Sai)$ is a monotone score
transformation but not a calibrated authorship probability.
Binary supervision is applied to $\Sfam$, and the log-sum-exp
aggregation can exceed the largest family margin by up to
$\tau_{\mathrm{agg}}\log G$ at each token. The measured expected
calibration error is $6.5\%$ under a balanced class prior.

\section{Training mixture and rewrite construction}
\label{app:mixture}

Table~\ref{tab:data} summarizes the training mixture by generator
family. The raw corpus is imbalanced, and the epoch sampling plan
reduces this imbalance. Rare families are oversampled toward a
target of $100{,}000$ draws, with total draws capped at four times
the number of available examples in that family. Large families
are capped at $1.1$ million draws. Additional
limits prevent any source, generator, or register from dominating
a family's sampled examples. These limits can leave a concentrated
family below the target, as for Claude in Table~\ref{tab:data}.
Within each length bucket, human examples are balanced across the
nine registers and make up half of each batch.

\begin{table}[tb]
\centering
\small
\setlength{\tabcolsep}{4pt}
\caption{\textbf{Training mixture by generator family.} All family tests
share the human examples. Public sources include RAID, MAGE, M4GT-Bench,
DetectRL, HC-Var, WildChat \citep{zhao2024wildchat}, and pre-2020 FineWeb
\citep{penedo2024fineweb}. Additional data include our newly constructed
paired arXiv corpus, peer reviews, and imitations of student writing. HC3, Ghostbuster, and
Beemo \citep{artemova2024beemo} are held out. Family names are abbreviated.}
\label{tab:data}
\begin{tabular}{@{}lrr>{\raggedright\arraybackslash}p{0.31\linewidth}r@{}}
\toprule
\textbf{Family test} & \textbf{Train rows} & \textbf{Val rows} &
\textbf{Examples of raw tags} &
\shortstack{\textbf{Epoch}\\\textbf{draws}} \\
\midrule
Human (shared) & $3{,}265{,}189$ & $20{,}019$ &
FineWeb, paired arXiv, reviews, essays & $50\%$/batch \\
\midrule
OpenAI GPT & $2{,}565{,}898$ & $2{,}877$ &
gpt2\,--\,gpt-4o, o1, gpt-oss-20b & $681$K \\
Other & $1{,}753{,}988$ & $7{,}796$ &
mpt, cohere, unattributed, evasion tools & $163$K \\
Mistral & $1{,}194{,}875$ & $4{,}441$ &
mistral, mistral-chat, ministral-3 & $788$K \\
Gemini/Gemma & $533{,}815$ & $6{,}079$ &
PaLM, gemini-3, gemma-4 & $780$K \\
Qwen & $491{,}549$ & $3{,}116$ &
qwen3.6-35b & $711$K \\
Llama & $428{,}172$ & $2{,}900$ &
llama\,1\,--\,3, llama-chat & $352$K \\
Claude & $65{,}756$ & $2{,}900$ &
Claude-instant, claude-sonnet-4.6 & $56$K \\
\midrule
Total & $10{,}299{,}242$ & $50{,}128$ &
$78$ raw generator tags & $7.06$M \\
\bottomrule
\end{tabular}
\end{table}

For the paired arXiv corpus, papers are divided into sections and
then into prose windows of approximately $600$ words. Equations
and citation keys are temporarily replaced by numbered placeholders,
such as \texttt{[MATH-1]} and \texttt{[CIT-1]}, before rewriting.
The original equations and citation keys are restored exactly afterward.
The nine operations range from
light editing (\emph{polish} and \emph{academic}) to larger changes
(\emph{paraphrase}, \emph{humanize}, \emph{roundtrip},
\emph{summarize}, and \emph{expand}). The remaining operations,
\emph{generate from context} and \emph{generate from abstract},
produce new section text using the paper's terminology.

Outputs must satisfy operation-specific length and copy-overlap
limits, as well as checks for markdown, meta-commentary,
repetition, and invented mathematics
(Appendix~\ref{app:rewrite}). After deduplication and removal of
short-window copies, $1{,}246{,}160$ source--rewrite pairs remain
in the training index.

\section{Evaluation and ablation protocol}
\label{app:protocol}

\paragraph{Surface-attribution ratio.}
SAR compares the median absolute process contrast with the median
absolute surface contrast, as defined in Section~\ref{sec:sar}.
It is invariant to nonzero linear rescaling of the score, but
not to arbitrary nonlinear transformations. No term in MELD's
training objective directly optimizes the process contrast
$\Delta_p$.

\paragraph{Evaluation protocol.}
Unless stated otherwise, each evaluation uses the full held-out
pool; MELD-eval is not subsampled. Thresholds for target
false-positive rates are computed from the human examples in
each pool. The reported operating points therefore measure
separability under pool-specific calibration, rather than the
transfer of a fixed threshold across populations. We emphasize
low-FPR operating points because AUROC is already close to its
maximum for several supervised detectors. Where reported,
$95\%$ percentile bootstrap confidence intervals use
$B=1{,}000$ resamples, and the $\pm$ values give their half-widths.
% AUTHOR CHECK: Specify the bootstrap resampling unit, treatment of
% paired/attacked examples, and whether thresholds are recomputed in
% each resample. The source does not provide these details.

\paragraph{Ablation protocol.}
The ablation protocol removes one component at a time from the
final configuration. The loss ablations remove family
cross-entropy, format consistency, or task cross-entropy.
Two data ablations compare removal of the pairing structure
with a volume-matched reduction that retains complete pairs.
Separate tests change the number of human anchors or the
pooling rule and rescore the trained checkpoint without
additional optimization. Appendix~\ref{app:stat_sig} describes
the comparisons and reports the results.

\section{Proofs}
\label{app:proofs}

\begin{proposition}[Bounded influence]
\label{prop:influence}
Let $v,v'\in[c^-,c^+]^n$ be fixed-length token-score vectors,
where $c^+-c^-\le 2\kappa$, and let
$q=\lceil\rho n\rceil$. If $v$ and $v'$ differ in at most
$m$ coordinates, then
\[
    \left|
        \operatorname{TQM}_\rho(v')
        -\operatorname{TQM}_\rho(v)
    \right|
    \le \frac{m(c^+-c^-)}{q}
    \le \frac{2\kappa m}{\lceil\rho n\rceil}.
\]
\end{proposition}

\begin{proof}
Equation~\eqref{eq:tokenlr} gives
$s_{\ell g}\in[-\kappa,\kappa]$ for every token and family.
The combined token score therefore lies in
\[
    c_\ell\in[c^-,c^+],
    \qquad
    c^\pm=\pm\kappa+\tau_{\mathrm{agg}}\log G.
\]
This interval has width $2\kappa$.

Changing one coordinate by an amount of absolute size $\Delta$
changes the mean of the $q$ largest coordinates by at most
$\Delta/q$. This holds whether the coordinate remains outside
the selected set, remains inside it, or crosses its boundary.
To move from $v$ to $v'$, change the differing coordinates one
at a time. There are at most $m$ such changes, each of size
at most $c^+-c^-$. Summing their effects gives
\[
    \left|
        \operatorname{TQM}_\rho(v')
        -\operatorname{TQM}_\rho(v)
    \right|
    \le\frac{m(c^+-c^-)}{q}
    \le\frac{2\kappa m}{\lceil\rho n\rceil}.
\]
\end{proof}

\begin{proposition}[Score stability under enrollment]
\label{prop:enroll}
Enrollment adds a new family center and bias while leaving all
other parameters fixed. Consequently, the human affinity
$\operatorname{hum}(u_\ell)$ and all existing family scores
$\Sfam[g]$, $g\le G$, remain unchanged on every input.
The combined token score can only increase, with
\[
\begin{aligned}
    c'_\ell-c_\ell
    &=\tau_{\mathrm{agg}}\log\!\left(
        1+\exp\!\left[
            \frac{s_{\ell,G+1}-c_\ell}{\tau_{\mathrm{agg}}}
        \right]
      \right) \\
    &\le\tau_{\mathrm{agg}}\exp\!\left[
        \frac{s_{\ell,G+1}-c_\ell}{\tau_{\mathrm{agg}}}
      \right].
\end{aligned}
\]
The document-score change satisfies
\[
    |\Sai'-\Sai|
    \le\max_\ell(c'_\ell-c_\ell).
\]
In particular, if the new family's clipped margin satisfies
$s_{\ell,G+1}\le c_\ell-\delta$ at every token of a document,
for $\delta>0$, then
\[
    |\Sai'-\Sai|
    \le\tau_{\mathrm{agg}}e^{-\delta/\tau_{\mathrm{agg}}}.
\]
\end{proposition}

\begin{proof}
The enrollment rule in Equation~\eqref{eq:enroll} appends
$(p_{G+1},b_{G+1})$ without changing the encoder, projection,
human anchors, or existing family parameters. Thus,
$u_\ell$, all human affinities, and all existing family
affinities remain unchanged. The same is true of every
existing token margin and pooled family score.

For the combined score, write
\[
    Z_\ell
    =\sum_{g\le G}e^{s_{\ell g}/\tau_{\mathrm{agg}}}
    =e^{c_\ell/\tau_{\mathrm{agg}}}.
\]
After enrollment,
\[
\begin{aligned}
    c'_\ell
    &=\tau_{\mathrm{agg}}\log\!\left(
        Z_\ell+e^{s_{\ell,G+1}/\tau_{\mathrm{agg}}}
      \right) \\
    &=c_\ell+\tau_{\mathrm{agg}}\log\!\left(
        1+e^{(s_{\ell,G+1}-c_\ell)/\tau_{\mathrm{agg}}}
      \right).
\end{aligned}
\]
The added term is nonnegative. Applying $\log(1+t)\le t$
for $t\ge0$ gives the stated bound on each token-score change.

Let $\Delta_{\max}=\max_\ell(c'_\ell-c_\ell)$. Every coordinate
increases by at most $\Delta_{\max}$. Top-quantile pooling is
monotone, and adding the same constant to all coordinates adds
that constant to the pooled score. It follows that
$|\Sai'-\Sai|\le\Delta_{\max}$. Under the token-wise condition
$s_{\ell,G+1}\le c_\ell-\delta$, we have
$\Delta_{\max}\le
\tau_{\mathrm{agg}}e^{-\delta/\tau_{\mathrm{agg}}}$.
\end{proof}

The enrollment adjustment in Appendix~\ref{app:enroll} is intended
to keep the new family's margins below the existing combined
scores on human text.

\section{New-family enrollment}
\label{app:enroll}

We enroll a new family using $N_{\mathrm{e}}=64$ documents from
its generator, with a minimum of eight documents. Each document
is encoded once, and its style representations are averaged over
valid tokens. Let $\mu$ be the mean of these document vectors,
and let $\varsigma$ be their mean distance from $\mu$.
The new center is
\begin{equation}
\begin{aligned}
    p_{G+1}
    &=(1-\lambda_{\mathrm{s}})
      \left[
        \mu+\alpha\varsigma
        \frac{\mu-a_{k^\ast}}{\lVert\mu-a_{k^\ast}\rVert}
      \right]
      +\lambda_{\mathrm{s}}\bar p, \\
    k^\ast
    &=\arg\min_k\lVert\mu-a_k\rVert,
\end{aligned}
\label{eq:enroll}
\end{equation}
where $\alpha=1.5$, $\lambda_{\mathrm{s}}=0.1$, and $\bar p$
is the mean of the existing family centers. The new bias is
set to the median existing family bias. We then save the
checkpoint with $G+1$ families.
% AUTHOR CHECK: Specify how the implementation handles mu = a_{k^*},
% where the normalized direction in this equation is undefined.

The adjustment moves the sample centroid away from its nearest
human anchor, with a displacement proportional to the sample
scatter, and then shrinks it toward the mean family center.
It is intended to address cases where the unadjusted centroid
lies near human representations and provides little separation.
The procedure requires no gradient updates, runs in seconds on
a single GPU, and preserves the human model and all existing
family tests (Proposition~\ref{prop:enroll}).

Related approaches address new generators through conformal
or group-specific thresholds \citep{zhu2025reliably,fairopt2025},
feature transfer with a frozen encoder \citep{ridgeft2026},
continual origin tracing \citep{glpd2025}, and few-shot
attribution of previously unseen image generators
\citep{omnidfa2025}.

\section{Training details}
\label{app:training}

Checkpoint selection uses a metric that averages overall AUROC,
worst-family AUROC, and both AUROC and fixed-threshold balanced
accuracy on the five weakest supported
family$\times$task$\times$register subsets. This metric increases
throughout the epoch, so we select the final checkpoint.
Human-score thresholds are computed from $22{,}190$ calibration
examples across nine registers and stored in the checkpoint.
Unless otherwise stated, evaluation uses offline fp32 scoring
with $2048$-token windows.

\paragraph{Batching and augmentation.}
We train on documents of up to $8192$ tokens using
length-bucketed batches with a fixed token budget. Computation
therefore depends on the length of each bucket rather than
the maximum document length. Human examples make up $50\%$
of each batch and provide the shared negatives in
Equation~\eqref{eq:lg}. Source--rewrite pairs are sampled as
individual examples rather than as paired training units.

Four data augmentations reduce formatting and minor editing
differences that the model could use to distinguish human from
AI training examples. For example, human arXiv passages often retain
hard line breaks from \TeX{} source, while AI rewrites usually do not.
With probability $0.3$, a document is replaced by a contiguous crop
of $48$--$192$ words; this operation is not applied to paired arXiv
examples. Hard line breaks are removed or reintroduced with
probability $0.75$ in both human and AI examples. Line wrapping
alone yields an AUROC of $0.74$, motivating this augmentation.

Markdown structures, including fences, headings, bullets, and
tables, are removed from examples that contain them with
probability $0.75$ and added to examples that lack them with
probability $0.25$. These rates are chosen to make markdown
similarly common in human and AI examples. A light post-editing
procedure introduces typos, omitted commas, and filler words
in AI examples with probability $0.25$. Its rate on human
examples is chosen to make these edits similarly common in both groups.

\paragraph{Optimization.}
The encoder and head are fine-tuned jointly with AdamW. We use
learning rates of $4.5\times10^{-5}$ for the encoder and
$6\times10^{-4}$ for the head, weight decay of $0.01$, and
$600$ warmup steps followed by cosine decay to $5\%$ of the
initial learning rate. Gradients are clipped at $1.0$.
Training uses bfloat16 mixed precision, with the scoring head
in fp32, for one epoch of the $7.06$-million-draw sampling
plan on eight L40 GPUs.

Anchors and centers are initialized from
$\mathcal{N}(0,0.5^2)$, and biases are initialized to zero.
The temperature is learned in log space, starting from
$\log\tau=-1$. During training, we periodically evaluate a
$12{,}000$-example validation subset with the deployment
scorer. We record AUROC and the formatting-sensitivity
measures from Section~\ref{sec:sar}: worst-format FPR,
FPR after format normalization, and SAR.

\section{Rewrite corpus}
\label{app:rewrite}

\paragraph{Source and units.}
We construct the corpus from a fixed revision of a public mirror
containing \LaTeX{} sources for $3{,}091{,}241$ arXiv papers.
Each paper is flattened, its macros are expanded, and its
sections are extracted. References, acknowledgments, and front
matter are discarded. Captions are retained as separate units,
and body prose is divided at sentence boundaries into windows
of approximately $600$ words, with a preferred range of
$400$--$850$ words and a hard limit of $1000$ words.
For rewriting, equations and citation keys are replaced by
temporary placeholders before generation and restored exactly afterward.

A unit is eligible only if every recorded version of its paper
predates 1~January 2021. We sample papers in round-robin order,
selecting at most three units per paper in each round. Captions
are capped at $5\%$ of the pool, and no arXiv category group
may contribute more than $38\%$. This yields $8.0$ million
units from $1{,}735{,}926$ papers. Train, validation, and test
assignments are made at the paper level, ensuring that the
splits are paper-disjoint and that both members of a
source--rewrite pair remain in the same split.

\paragraph{Generators.}
Four open-weight instruction models produce the corpus:
Gemma-4-26B-A4B-it at temperature $0.65$,
Qwen3.6-35B-A3B at $0.70$ with thinking disabled,
Ministral-3-14B-Instruct-2512 at $0.55$, and
gpt-oss-20b at $0.70$. All use top-$p$ sampling with
$p=0.95$ and a maximum output length of $2400$ tokens.
Each source unit is assigned to one generator. Units with
the greatest mathematical content are restricted to
\emph{polish} and \emph{academic}, the two light-editing
operations.

All four models receive the same system instruction, which
prohibits preambles, markdown, lists, and statements about
being an AI or about the changes made. Rejected outputs are
retried at a slightly higher temperature, with a hint
identifying the failed acceptance check. Rewrites from all
four generators enter training.

\paragraph{Held-out generator test set.}
The unseen-family evaluation in Section~\ref{sec:val} uses
the same construction pipeline. DeepSeek-V4-Flash and
DeepSeek-V4-Pro, both with thinking disabled, rewrite the
human passages from gpt-oss-20b's held-out test shard.
The prompts, placeholder replacement procedure, and acceptance checks
are unchanged. The resulting test set contains $1{,}600$ pairs
per model across the nine operations. Neither DeepSeek
model contributes training examples. Because the source
passages are shared, the gpt-oss-20b subset provides a
matched control for a generator represented in training.

\begin{table}[ht]
\centering
\small
\setlength{\tabcolsep}{4pt}
\caption{\textbf{Rewrite operations and acceptance criteria.} The nine
operations map to six training task labels. \emph{Length} gives the
allowed ratio of output words to source words, except for the two
generation operations, which use absolute word-count limits because
they have no source passage. \emph{Copy} is the maximum allowed
$5$-gram containment of the source in the output. Polish and academic
allow greater overlap because they make smaller edits. \emph{Rows}
counts AI examples in the assembled corpus.}
\label{tab:ops}
\begin{tabular}{@{}
    >{\raggedright\arraybackslash}p{0.22\linewidth}
    >{\raggedright\arraybackslash}p{0.37\linewidth}
    r r r@{}}
\toprule
\textbf{Operation} & \textbf{What the instruction asks for} &
\textbf{Length} & \textbf{Copy} & \textbf{Rows} \\
\midrule
Polish &
Light copy-editing while preserving the author's voice,
terminology, and length &
$0.68$--$1.35$ & $0.90$ & $668$K \\

Paraphrase &
Change wording and sentence structure while preserving
every claim, number, and logical step &
$0.62$--$1.40$ & $0.68$ & $203$K \\

Academic &
Use a formal journal style with precise terminology and
appropriate hedging; preserve content &
$0.62$--$1.40$ & $0.90$ & $178$K \\

Humanize &
Use a natural, slightly informal first-person academic voice;
preserve technical content &
$0.62$--$1.40$ & $0.68$ & $131$K \\

Roundtrip &
Rewrite as though translated into another language and then
back into English by a fluent scientist &
$0.62$--$1.40$ & $0.68$ & $121$K \\

Summarize &
Reduce the passage to about half its length while preserving
the main claims and reasoning &
$0.30$--$0.85$ & $0.68$ & $98$K \\

Expand &
Expand to approximately $1.35\times$ the source length by
explaining implicit steps, without adding unsupported content &
$1.08$--$2.00$ & $0.68$ & $78$K \\

Generate from context &
Write a new section from the title, abstract, and excerpts
from the paper's other sections &
$130$--$900$\,w & $0.68$ & $65$K \\

Generate from abstract &
Write a new section from the title, abstract, and section
heading alone &
$130$--$900$\,w & $0.68$ & $50$K \\
\bottomrule
\end{tabular}
\end{table}

\paragraph{Acceptance, pairing, and assembly.}
Table~\ref{tab:ops} gives the operation-specific length and
copy-overlap limits. We additionally reject outputs that
contain fewer than $15$ words, markdown, meta-commentary,
emoji, excessive repetition, or invented mathematics. Outputs that
omit a placeholder or add one that was not in the source are also
rejected. Near-verbatim outputs
are excluded to avoid assigning an AI label to a passage
that is effectively unchanged from its human source.
This check removes $22.6\%$ of gpt-oss-20b outputs,
$10.8\%$ of Gemma outputs, $11.0\%$ of Qwen outputs,
and $0.1\%$ of Ministral outputs.

We use approximately $400{,}000$ outputs per model
($391{,}000$ for Qwen) and produce $3{,}068{,}271$ examples.
These comprise $1{,}476{,}558$ source--rewrite pairs, each
containing one human and one AI version of a section, and
approximately $115{,}000$ unpaired examples from the two
\emph{generate} operations. The latter produce new section
text rather than rewrites of a source passage and are
therefore included as ordinary AI examples. Equations and
citation keys are preserved exactly for the seven rewriting
operations, but not for the two generation operations.

Both members of each pair undergo format normalization during corpus
construction. Without this step, the human source retains
\TeX{} line breaks while the model output is usually
reflowed. Newline density alone distinguishes the two
members in $93\%$ of pairs; with the line-wrap augmentation
in Appendix~\ref{app:training}, this falls to $56\%$.
We do not normalize word length, which differs between the
two classes ($5.83$ versus $6.09$ characters on average).

The nine operations map to six training labels:
\emph{polish} and \emph{academic} map to polish;
\emph{paraphrase} and \emph{roundtrip} map to paraphrase;
and both generation operations map to generate. Humanize,
summarize, and expand retain their own labels. Every arXiv
identifier used in training is recorded, and all contributing
papers are excluded from the case study in
Section~\ref{sec:study}.

\section{MELD-eval}
\label{app:meld_eval}

MELD-eval evaluates transfer to four held-out generators under a common domain and attack protocol.
It reuses RAID-style English domains, human seeds, and
attacks to reduce differences unrelated to the generators.

\paragraph{Generators and domains.}
The pool covers eight English domains using two prompting
modes. In \emph{continuation} mode, used for books, news,
and reddit, the model extends a short human seed. In
\emph{instruction} mode, used for abstracts, recipes,
reviews, wiki, and poetry, the model generates text from
a topic line. Table~\ref{tab:meld_eval_seeds} lists the
per-generator example counts, domain-specific seed limits,
target lengths, and prompt inputs.

\begin{table}[ht]
\centering
\small
\caption{\textbf{MELD-eval construction.} \emph{Top:} clean and
attacked AI examples per generator, with each generator's share of all
AI examples. The $7{,}862$ human seeds are shared across generators.
\emph{Bottom:} domain-specific prompting settings. Mode A continues a
human seed; Mode B generates text from a topic line. The reviews domain
has $862$ seeds after English-language filtering.}
\label{tab:meld_eval_seeds}

\setlength{\tabcolsep}{8pt}
\begin{tabular}{lrrr}
\toprule
\multicolumn{4}{l}{\textit{Per-generator counts}} \\
\textbf{Generator} & \textbf{Clean AI} &
\textbf{Attacked AI} & \textbf{Ratio} \\
\midrule
GPT-5.4-Mini (OpenAI)        & $7.86$K & $47.17$K & $0.25$ \\
Gemini-3-Flash (Google)      & $7.86$K & $47.17$K & $0.25$ \\
Claude-Haiku-4.5 (Anthropic) & $7.86$K & $47.17$K & $0.25$ \\
Qwen-3.6-Plus (Alibaba)      & $7.86$K & $47.17$K & $0.25$ \\
\midrule
\textit{Total AI} & $31.45$K & $188.69$K & $1.00$ \\
\bottomrule
\end{tabular}

\vspace{0.6em}

\setlength{\tabcolsep}{8pt}
\begin{tabular}{llrrl}
\toprule
\multicolumn{5}{l}{\textit{Per-domain configuration}} \\
\textbf{Domain} & \textbf{Mode} & \textbf{Seed cap} &
\textbf{Target words} & \textbf{Seed input} \\
\midrule
books     & A & $1{,}000$ & $500$ & first $80$ words of human seed \\
news      & A & $1{,}000$ & $500$ & first $80$ words of human seed \\
reddit    & A & $1{,}000$ & $300$ & first $60$ words of human seed \\
abstracts & B & $1{,}000$ & $250$ & topic line \\
recipes   & B & $1{,}000$ & $400$ & topic line \\
reviews   & B & $862$     & $400$ & topic line \\
wiki      & B & $1{,}000$ & $500$ & topic line \\
poetry    & B & $1{,}000$ & $180$ & topic line \\
\bottomrule
\end{tabular}
\end{table}

\paragraph{Prompting and decoding.}
All generators receive a shared system instruction requesting
plain text without preambles, meta-commentary, chain-of-thought,
or markdown. Continuation prompts ask the model to maintain
the seed's voice and register. Instruction prompts provide
only a topic line and request text in the target domain.
We use temperature $0.7$, top-$p$ $0.95$, and an output
limit of $1024$ tokens. Qwen-3.6-Plus is queried with
thinking disabled to keep its outputs comparable in length
and style. The code release contains the exact prompt
templates and model identifiers.

\paragraph{Attacks and text normalization.}
Each of the $31{,}448$ clean AI examples receives six
RAID-style attacks: homoglyph substitution, number
substitution, synonym substitution, case changes,
whitespace modification, and zero-width-space insertion.
We use RAID's per-token attack rates, producing
$188{,}688$ attacked examples. For reviews, recipes, and
poetry, residual markdown is removed from both AI outputs
and paired human text using the same procedure, reducing
formatting differences as a detection cue.

\section{Qualitative examples}
\label{app:cases}

Table~\ref{tab:qualitative} compares MELD, Binoculars,
ModernBERT-Detect, and RoBERTa-ChatGPT on selected cases
where their predictions disagree. The examples come from
HC3, DetectRL-test, and MELD-eval, illustrating clean
classification, robustness to attacks, and transfer to
held-out generators, respectively.

\input{figures/tab_qualitative}

In the selected HC3 cases, baselines make errors at the
chosen operating point despite their high aggregate AUROC,
while MELD classifies all shown examples correctly. In
DetectRL, MELD detects the displayed attacked AI examples
and avoids the false positive that ModernBERT-Detect makes
on the paraphrased human example. In MELD-eval, MELD
detects the displayed clean and attacked examples from
all four generators, whereas the baselines miss most of
them. These selected cases illustrate individual errors;
aggregate performance is reported in
Tables~\ref{tab:bench} and~\ref{tab:meld_eval}.

\section{Ablations and attribution results}
\label{app:stat_sig}

\paragraph{Unseen-family and additional benchmarks.}
The DeepSeek-V4-Flash and DeepSeek-V4-Pro test sets use
held-out arXiv passages and the rewrite pipeline described
in Appendix~\ref{app:rewrite}. No DeepSeek family is present
in training. The family head assigns most DeepSeek examples
to OpenAI GPT ($67$--$82\%$) or Qwen ($15$--$26\%$), often
with high margins. Requiring an attribution margin above
$2$ excludes $22$--$29\%$ of these examples.

Beemo remains challenging for all evaluated detectors:
ModernBERT-Detect, RoBERTa-ChatGPT, and RoBERTa-Large achieve
AUROCs of $51$--$69$, compared with approximately $82$ for
MELD. One possible explanation is the labeling convention
under which lightly edited AI text can appear on the human
side of the evaluation. On TuringBench, which contains
2019-era generated text, MELD achieves an AUROC of $80.1$
\citep{uchendu2021turingbench}.

\paragraph{Single-component ablations.}
Each loss ablation removes one component from the full
configuration while keeping the corpus, seed, optimization
schedule, and token budget fixed. The evaluation protocol
uses the same $12{,}000$ validation examples, $186$ documents
for formatting tests, and $150$ held-out source--rewrite
pairs for each configuration.

Two data ablations distinguish the effect of pairing from
the effect of corpus size. The \emph{unpaired} configuration
removes one member of every pair, so a source passage no
longer appears alongside its rewrite under the opposite
label. The \emph{half-pairs} control removes complete pairs
while retaining the pairing structure in the remaining
examples. Both remove approximately $12\%$ of the full
corpus. Their comparison is intended to measure the
contribution of retaining matched examples beyond the
amount of training data.

\begin{table}[ht]
\centering
\caption{\textbf{Single-component ablation results.} The first panel reports AUROC, TPR at $1\%$ FPR, family
attribution accuracy, and coverage and accuracy at attribution margin
$>1$. The second reports raw worst-format FPR, FPR after format normalization,
SAR, and task accuracy. AUROC, TPR, accuracy, and coverage are on a
$0$--$100$ scale. SAR uses the training-time evaluation pairs, which
differ from those in Table~\ref{tab:sar}.
``---'' indicates that no task head is trained.}
\label{tab:ablation_v7}
\setlength{\tabcolsep}{4pt}
\small
\begin{tabular}{@{}lrrrr@{}}
\toprule
\textbf{Configuration} & \textbf{AUROC} &
\textbf{TPR@1\%} & \textbf{Family acc.} &
\shortstack{\textbf{Coverage}\\\textbf{@ accuracy}} \\
\midrule
MELD (full) &
$99.80$ & $98.25$ & $90.9$ & $88.9$ @ $95.4$ \\
no family CE &
$99.52$ & $96.10$ & $62.6$ & $8.8$ @ $92.2$ \\
no format consistency &
$99.61$ & $96.84$ & $89.7$ & $86.2$ @ $94.8$ \\
no task CE &
$99.70$ & $97.45$ & $89.9$ & $87.5$ @ $95.0$ \\
unpaired corpus &
$99.23$ & $94.62$ & $88.4$ & $85.1$ @ $94.1$ \\
half pairs (volume control) &
$99.58$ & $96.71$ & $89.8$ & $87.3$ @ $94.9$ \\
\bottomrule
\end{tabular}

\medskip
\begin{tabular}{@{}lrrrr@{}}
\toprule
\textbf{Configuration} & \textbf{Raw FPR} &
\shortstack{\textbf{Normalized}\\\textbf{FPR}} &
\textbf{SAR} & \textbf{Task acc.} \\
\midrule
MELD (full) &
$1.61\%$ & $0.54\%$ & $144.8$ & $95.6$ \\
no family CE &
$2.15\%$ & $0.54\%$ & $121.9$ & $94.9$ \\
no format consistency &
$8.60\%$ & $0.54\%$ & $57.8$ & $95.1$ \\
no task CE &
$1.61\%$ & $0.54\%$ & $131.2$ & --- \\
unpaired corpus &
$2.15\%$ & $0.54\%$ & $88.4$ & $92.7$ \\
half pairs (volume control) &
$2.15\%$ & $0.54\%$ & $118.6$ & $93.8$ \\
\bottomrule
\end{tabular}
\end{table}

We also examine the scoring geometry by modifying the trained
head and rescoring without additional optimization. Replacing
eight human anchors with one raises the median score on the
human calibration pool from $-5.5$ to $+2.2$. The proportion
of these documents with scores above zero rises from $0.8\%$
to $100\%$. Replacing top-quantile pooling with mean pooling
reduces AUROC by $1.3$ points on documents containing
$10\%$ AI text.

\paragraph{Per-family detection.}
Table~\ref{tab:family_auroc} evaluates each family test on
its own AI examples against the shared human validation
examples. AUROC ranges from $99.68$ to $99.91$, a spread
of $0.23$ points. Qwen and the heterogeneous Other family
have the lowest AUROC, matching the worst-family value
reported in Section~\ref{sec:val}.

\begin{table}[ht]
\centering
\small
\setlength{\tabcolsep}{6pt}
\caption{\textbf{Per-family detection AUROC on validation.}
Each family test is evaluated on that family's AI examples against
the shared human examples. AUROC is reported on a $0$--$100$ scale.}
\label{tab:family_auroc}
\begin{tabular}{lrr}
\toprule
\textbf{Family test} & \textbf{AI rows} & \textbf{AUROC} \\
\midrule
Mistral      & $4{,}441$ & $99.91$ \\
Claude       & $2{,}900$ & $99.87$ \\
OpenAI GPT   & $2{,}877$ & $99.81$ \\
Llama        & $2{,}900$ & $99.80$ \\
Gemini/Gemma & $6{,}079$ & $99.80$ \\
Other        & $7{,}796$ & $99.68$ \\
Qwen         & $3{,}116$ & $99.68$ \\
\bottomrule
\end{tabular}
\end{table}

\paragraph{Attribution accuracy.}
We evaluate the family and task heads separately from binary
detection. The six-way task head achieves $95.6\%$ top-1
accuracy on validation AI examples. Family attribution
achieves $91.9\%$ on the validation sample described
in Section~\ref{sec:val}; Table~\ref{tab:heads} gives the
per-family precision and recall. Claude has $99.2\%$ recall
and $99.7\%$ precision, while the heterogeneous Other family
has the lowest recall at $82.0\%$.

The attribution-margin threshold trades coverage for accuracy.
At a margin greater than $1$, the detector retains $90.1\%$
of examples with $96.1\%$ top-1 accuracy. Raising the threshold
to $2$ retains $81.4\%$ of examples with $98.3\%$ accuracy.

\begin{table}[ht]
\centering
\small
\setlength{\tabcolsep}{5pt}
\caption{\textbf{Task and family attribution.} Top-1 accuracy is
reported for the six-way task head and seven-way family head.
Family precision and recall use validation examples sampled by writing
category, with $1{,}200$ AI examples per family. All values are
percentages. ``---'' indicates that only top-1 accuracy is reported
for the task head.}
\label{tab:heads}
\begin{tabular}{lrrr}
\toprule
\textbf{Prediction} & \textbf{Top-1 acc.} &
\textbf{Macro prec.} & \textbf{Macro recall} \\
\midrule
task ($6$-way)   & $95.6$ & --- & --- \\
family ($7$-way) & $91.9$ & $92.0$ & $91.9$ \\
\midrule
\multicolumn{2}{l}{\textit{Per family}} &
\textit{Precision} & \textit{Recall} \\
\quad Claude       & & $99.7$ & $99.2$ \\
\quad Llama        & & $93.3$ & $95.6$ \\
\quad Mistral      & & $87.4$ & $92.6$ \\
\quad Qwen         & & $95.5$ & $92.4$ \\
\quad OpenAI GPT   & & $86.5$ & $91.8$ \\
\quad Gemini/Gemma & & $91.0$ & $89.8$ \\
\quad Other        & & $90.7$ & $82.0$ \\
\bottomrule
\end{tabular}
\end{table}

\section{Case-study methods}
\label{app:study_details}

Previous studies have used distributional estimators to
measure LLM-modified content in peer reviews
\citep{liang2024monitoring} and scientific abstracts
\citep{liang2024mapping}. Other work has reported an
association between AI-assisted reviews and higher scores
at one venue \citep{latona2024ai}.
Our case study in Section~\ref{sec:study} analyzes eligible
arXiv papers, abstracts from nine conferences, and peer
reviews at section and monthly resolution. The analysis
uses training-overlap controls, matched review fields,
baseline-year negative controls, and threshold-free
percentiles.

Each group has a separate threshold corresponding to a
nominal $1\%$ exceedance rate in its baseline population.
The observed rate ranges from $0.81\%$ to $1.47\%$ across
the seven baseline years, 2016--2022, which serve as negative
controls. Threshold-free percentiles are the primary
statistics.

\begin{table}[tb]
\centering
\small
\setlength{\tabcolsep}{3.5pt}
\caption{\textbf{arXiv detector scores, 2016--2026.} Share above the $1\%$ baseline threshold; median section's role-specific baseline percentile and shift in baseline SD. $^{*}$2026: January--July.}
\label{tab:study_year}
\begin{tabular}{@{}lrrrrrrrrrrr@{}}
\toprule
\textbf{Year (20xx)} & \textbf{16} & \textbf{17} & \textbf{18} & \textbf{19} & \textbf{20} & \textbf{21} & \textbf{22} & \textbf{23} & \textbf{24} & \textbf{25} & \textbf{26}$^{*}$ \\
\midrule
Flagged (\%)        & $1.3$ & $1.5$ & $1.0$ & $1.0$ & $0.8$ & $0.8$ & $0.9$ & $4.1$ & $10.5$ & $23.0$ & $25.4$ \\
Median pctile (\%)  & $47.7$ & $49.0$ & $48.4$ & $49.1$ & $50.9$ & $52.7$ & $54.3$ & $60.9$ & $71.0$ & $85.2$ & $89.5$ \\
Shift (baseline SD)  & $0.00$ & $0.00$ & $0.00$ & $0.00$ & $0.00$ & $0.00$ & $+0.01$ & $+0.02$ & $+0.06$ & $+0.35$ & $+0.72$ \\
Section units (K)   & $13$ & $17$ & $26$ & $56$ & $186$ & $727$ & $759$ & $868$ & $1{,}034$ & $1{,}233$ & $762$ \\
\bottomrule
\end{tabular}
\end{table}

\paragraph{Sections and authors.}
Table~\ref{tab:study_sections} reports changes by section
role and primary arXiv category using
$P(\text{more AI-like})$, defined as the probability that
a text from a given year scores above a matched baseline
text. This mean-rank statistic is lower than the median
percentile in Table~\ref{tab:study_year}, reflecting the
portion of sections that remain at low detector-score
percentiles. To assess changes within authors, we also
compare $23{,}404$ first authors with at least three publications
before 2023 and at least three from 2023 onward. Of these authors, $77\%$ have higher later score
percentiles, with a median change of $+10$ percentile points.

\paragraph{Peer review and outcomes.}
Because review forms change over time, the primary review
series uses the summary field whose form is consistent
across the included venue--year groups. In 2025, $38.6\%$
of ICLR review summaries exceed their baseline threshold,
compared with $23.0\%$ of arXiv section units. Analyses of
ratings and acceptance describe population-level
associations rather than causal effects.

\paragraph{Corpora.}
The arXiv corpus contains all $1{,}344{,}610$ eligible papers
from 2016--2026, divided into $5{,}682{,}241$ units with
seven section roles: abstract, introduction, related work,
methods, results, discussion, and conclusion. Eligibility
excludes all $1{,}700{,}807$ paper identifiers recorded as
contributing to MELD's training pool.

The conference corpus contains $92{,}174$ accepted abstracts
from CVPR, ICCV, ECCV, NeurIPS, ICML, ICLR, ACL, EMNLP,
and NAACL, covering 2016--2025. The primary peer-review
series contains $143{,}685$ form-matched summaries, with
reported lengths of $94$--$104$ words across venue--year
groups. A secondary series contains $170{,}720$ full-prose
reviews. Outcome analyses use $19{,}064$ ICLR submissions
with review ratings and acceptance decisions.
% AUTHOR CHECK: Specify the length summary represented by the
% 94--104-word range; the source does not identify mean vs. median.

The most recent available data was until 11~July 2026. Accordingly,
2026 is a partial year throughout the analysis. Baseline
references are balanced across years from 2016--2022.
The measured baseline false-positive rate is $1.05\%$
for arXiv and $1.02\%$ for reviews.

\paragraph{Detector and reference distributions.}
The study uses a MELD checkpoint frozen before scoring,
with a nine-operation task head matching the generation
pipeline and the labels in Figure~\ref{fig:study_transition}b
(Appendix~\ref{app:rewrite}). We report
$P(\text{AI})=\sigma(\Sai)$ only as a monotone transformation
of the detector score, not as a calibrated probability.
All primary statistics are rank-based and therefore
unchanged by this transformation. The reference distributions
contain $1{,}784{,}201$ arXiv section units, $409{,}743$
arXiv paper abstracts, and $74{,}875$ review units from
years through 2022.

\paragraph{Training-overlap control.}
The training pool records arXiv identifiers for scientific-text
examples. We exclude all $1{,}700{,}807$ recorded paper IDs
before selecting the study corpus. The final overlap between
the scored corpus and the training set is zero among the
$1{,}344{,}610$ eligible papers.

\paragraph{Additional controls.}
\emph{Leave-one-year-out calibration.}
For each baseline year, we recompute the threshold without
that year and evaluate the held-out examples. The resulting
flagged rates range from $0.86\%$ to $1.64\%$.

\emph{Equal sample sizes.}
Sampling $12{,}769$ units from every year, equal to the size
of the smallest annual group, reproduces the temporal pattern.

\emph{Paper versions.}
Restricting the analysis to first submissions (v1) gives a
similar series. The flagged share in partial-year 2026 is
$25.1\%$, compared with $25.4\%$ in the full corpus.
The pattern therefore persists after controlling for
revision status in the frozen mirror.

\emph{Review length.}
The 2024--2025 increase remains within every review-length
bucket, from $40$--$149$ words to $900$ or more words.

\emph{Review forms.}
The change in the ICLR review form between 2021 and 2022,
both baseline years, produces score shifts whose direction
and magnitude vary across detector checkpoints. We therefore
compare only matched review fields across years. As a
separate negative control, NeurIPS~2021 reviews are scored
against the 2022 reference and have a median percentile
of $48.5\%$, close to the baseline value of $50\%$.

\emph{Thresholds and corpus size.}
The temporal pattern persists at baseline exceedance levels
of $1\%$, $5\%$, and $10\%$, and under the threshold-free
statistics. Expanding the analysis from a $613{,}000$-paper
sample to the full eligible corpus changes no annual
median percentile by more than $0.1$ points.

\begin{figure}[t]
\centering
\includegraphics[width=0.55\textwidth]{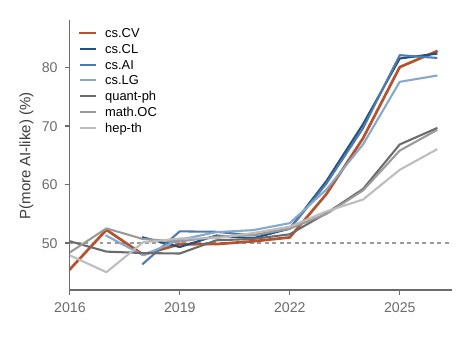}
\caption{\textbf{Detector-score trends by primary arXiv category,
2016--2026.} Each curve gives the probability that a text from a given
year scores above a 2016--2022 baseline text from the same category.
The reference level is $50\%$.}
\label{fig:study_where}
\end{figure}

\paragraph{Fields, venues, and replication.}
Score changes differ across primary arXiv categories
(Table~\ref{tab:study_sections} and
Figure~\ref{fig:study_where}). In partial-year 2026,
cs.CV reaches $P(\text{more AI-like})=82.7\%$, with a
shift of approximately $+3.6$ baseline standard deviations.
For hep-th, the corresponding values are $66.0\%$ and
$+0.08$ standard deviations. In cs.AI, the monthly flagged
share falls from $47\%$ in January to $30\%$ in June 2026,
alongside a change in the energy score. A possible
explanation is a change in generator writing styles.

With separate baseline references, the corresponding
rank-based values for bioRxiv and medRxiv abstracts reach
$77\%$ and $85\%$ in 2026. For accepted conference abstracts
in 2025, venue-specific baseline percentiles range from
the $89$th to the $98$th percentile. Flags also cluster
within papers, with a conditional lift of $1.9\times$
in 2026. For papers with at least three scored units, this is
the probability that another unit is flagged given a flagged
unit, divided by the mean within-paper flagged fraction.

\begin{figure}[t]
\centering
\includegraphics[width=\textwidth]{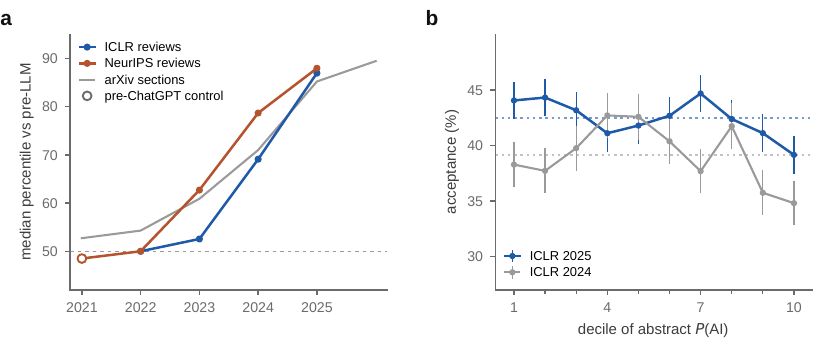}
\caption{\textbf{Review-score trends and ICLR acceptance rates.}
\emph{(a)} Form-matched review summaries, with each venue compared
against its own last baseline year. The concurrent arXiv series is
shown for reference, and open markers denote negative controls.
\emph{(b)} Acceptance rates by within-year decile of abstract
$P(\text{AI})$. Error bars show one standard error of the mean;
dotted lines show the annual mean. Acceptance is approximately
constant across the first nine deciles and lower in the highest
decile.}
\label{fig:study_reviews}
\end{figure}

\paragraph{Within-author analysis.}
For each of the $23{,}404$ first authors, we compare section
roles present in both periods using the same baseline
reference. Changes form a continuous distribution.
Using the study's change categories, $36.4\%$ remain
within baseline variation, $41.8\%$ show modest changes,
$20.2\%$ show substantial changes, and $1.6\%$ fall in
the largest-change category.
% AUTHOR CHECK: Supply the thresholds defining these change categories
% and the dates defining the two comparison periods.

\begin{table}[tb]
\centering
\small
\setlength{\tabcolsep}{6pt}
\caption{\textbf{Score shifts by section role and arXiv category
in 2026.} Each group is compared with its own baseline
distribution. \emph{Top:} section roles. Conclusions show the largest
standardized shift ($+3.39$ baseline standard deviations).
\emph{Bottom:} selected primary categories. The listed computer-science
categories show larger shifts than the listed physics and mathematics
categories.}
\label{tab:study_sections}
\begin{tabular}{lrr}
\toprule
\textbf{Section role} &
\textbf{P(more AI-like) (\%), 2026} &
\textbf{Shift (baseline SD)} \\
\midrule
conclusion   & $81.4$ & $+3.39$ \\
discussion   & $80.1$ & $+2.03$ \\
related work & $78.8$ & $+1.56$ \\
introduction & $76.8$ & $+0.62$ \\
methods      & $74.9$ & $+0.68$ \\
results      & $74.9$ & $+0.89$ \\
abstract     & $73.5$ & $+0.92$ \\
\bottomrule
\end{tabular}

\par\vspace{0.8em}

\begin{tabular}{lrr}
\toprule
\textbf{Category} &
\textbf{P(more AI-like) (\%), 2026} &
\textbf{Shift (baseline SD)} \\
\midrule
cs.CV    & $82.7$ & $+3.63$ \\
cs.CL    & $82.4$ & $+2.70$ \\
cs.AI    & $81.6$ & $+2.89$ \\
cs.LG    & $78.6$ & $+1.65$ \\
quant-ph & $69.6$ & $+0.20$ \\
math.OC  & $69.3$ & $+0.12$ \\
hep-th   & $66.0$ & $+0.08$ \\
\bottomrule
\end{tabular}
\end{table}

\begin{table}[tb]
\centering
\small
\setlength{\tabcolsep}{6pt}
\caption{\textbf{Form-matched review summaries by venue and year.}
Each venue uses its own last baseline year as the reference.
NeurIPS~2021 is a negative control evaluated against the 2022
reference; its median percentile is close to the reference value
of $50\%$.}
\label{tab:study_reviews}
\begin{tabular}{llrrrr}
\toprule
\textbf{Venue} & \textbf{Role} & \textbf{Year} &
\textbf{\textit{n}} & \textbf{Median pctile (\%)} &
\textbf{Shift (baseline SD)} \\
\midrule
NeurIPS & control  & 2021 & $9{,}676$  & $48.5$ & $-0.03$ \\
NeurIPS & baseline & 2022 & $9{,}536$  & $50.0$ & $\phantom{-}0.00$ \\
NeurIPS &          & 2023 & $14{,}060$ & $62.7$ & $+0.39$ \\
NeurIPS &          & 2024 & $15{,}300$ & $78.7$ & $+1.26$ \\
NeurIPS &          & 2025 & $21{,}278$ & $88.0$ & $+1.97$ \\
\midrule
ICLR    & baseline & 2022 & $9{,}058$  & $50.0$ & $\phantom{-}0.00$ \\
ICLR    &          & 2023 & $12{,}709$ & $52.6$ & $+0.06$ \\
ICLR    &          & 2024 & $20{,}069$ & $69.1$ & $+0.66$ \\
ICLR    &          & 2025 & $31{,}999$ & $87.0$ & $+1.86$ \\
\bottomrule
\end{tabular}
\end{table}

\paragraph{Rating analyses.}
Table~\ref{tab:study_reviews} and
Figure~\ref{fig:study_reviews}a give the full form-matched
review series. Rating scales differ across venues and
years: ICLR~2020 uses $\{1,3,6,8\}$, and average NeurIPS
ratings are approximately one point higher than ICLR
ratings. Before making comparisons, we therefore convert
each rating to its percentile within the same venue
and year.

Flagged and unflagged reviews have similar calibrated
ratings. The differences are $-0.7$ percentile points
for ICLR~2025 and $+0.5$ points for ICLR~2024.
Within calibrated-rating quintiles, the flagged share of
a paper's reviews is associated with acceptance-rate
differences ranging from $-0.8$ to $+5.5$ percentage
points, with no consistent direction.

Papers with the highest abstract detector scores receive
lower calibrated ratings overall. For ICLR 2024 and 2025,
the reported differences are $-6.8$ and $-5.6$
percentile points, with $z=-4.5$ and $z=-7.9$,
respectively. However, acceptance rates are approximately
constant across the first nine abstract-score deciles
and lowest in the highest decile
(Figure~\ref{fig:study_reviews}b). The top-decile acceptance
rates are $34.8\%$ and $39.2\%$ in 2024 and 2025, respectively,
compared with corresponding mid-range rates of
approximately $40\%$ and $43\%$.
Before 2024, flagged reviews tend to be unusually short.
Flagged and unflagged reviews have similar median lengths:
about $374$ versus $371$ words in 2024 and $384$ versus $391$
in 2025.
The later comparisons therefore do not show the same
length difference as the earlier flagged reviews.

\end{document}

%% file: figures/main_table.tex
\begin{tabular*}{\textwidth}{@{\extracolsep{\fill}}lrrrrrr@{}}
\toprule
 & \multicolumn{3}{c}{\textbf{All settings (with attack)}} & \multicolumn{3}{c}{\textbf{No attack}} \\
\cmidrule(lr){2-4}\cmidrule(lr){5-7}
\textbf{Detector} & \textbf{AUROC} & \textbf{TPR@$5\%$} & \textbf{TPR@$1\%$} & \textbf{AUROC} & \textbf{TPR@$5\%$} & \textbf{TPR@$1\%$} \\
\midrule
\multicolumn{7}{l}{\emph{\textbf{Commercial detectors (closed-source)}}} \\
\citet{grammarly_detector}                      & \textbf{99.87}    & 99.47             & \underline{98.21} & \textbf{99.97}    & \textbf{99.91}    & \textbf{99.69}    \\
\citet{quillbot_detector}                        & 99.62             & \underline{99.62} & 98.10             & 99.78             & 99.72             & 99.24             \\
GPTZero \citep{adam2026gptzero}     & 99.25             & 97.24             & 92.93             & 99.53             & 98.37             & 95.72             \\
\midrule
\multicolumn{7}{l}{\emph{\textbf{Open-source detectors}}} \\
ModernBERT-Detect     & 97.65 & 94.14 & 88.23 & 99.12 & 99.11 & 98.18 \\
Binoculars                       & ---   & ---   & ---   & 84.40 & 78.98 & 69.54 \\
RADAR                                 & 81.92 & 63.91 & 43.12 & 81.89 & 65.61 & 48.09 \\
GLTR                             & 70.90 & 51.48 & 36.48 & 72.68 & 59.69 & 45.57 \\
LLMDet \citep{wu2023llmdet}         & 62.90 & 26.70 & 14.91 & 65.92 & 33.40 & 20.16 \\
\midrule
\textbf{MELD} (ours)                                      & \underline{99.82} & \textbf{99.78}    & \textbf{99.24}    & \underline{99.85} & \underline{99.76} & \underline{99.40} \\
\bottomrule
\end{tabular*}

%% file: figures/tab_benchmarks.tex
\begin{tabular}{@{}lrrrrrr@{}}
\toprule
\textbf{Detector} & \textbf{HC3} & \textbf{MAGE} & \textbf{M4GT} & \textbf{Ghostbuster} & \textbf{DetectRL} & \textbf{MELD-eval} \\
\midrule
\multicolumn{7}{l}{\emph{Zero-shot detectors (re-evaluated on the same pools as MELD)}} \\
GLTR & \tableci{99.5}{0.1} & \tableci{64.4}{0.5} & \tableci{73.4}{1.7} & \tableci{88.2}{1.0} & \tableci{71.4}{0.5} & \tableci{42.4}{0.6} \\
Fast-DetectGPT & \tableci{99.1}{0.1} & \tableci{57.1}{0.5} & \tableci{65.9}{1.5} & \tableci{92.6}{0.8} & \tableci{73.0}{0.5} & \tableci{70.5}{0.4} \\
Binoculars & \tableci{79.4}{0.4} & \tableci{60.7}{0.5} & \tableci{57.3}{1.8} & \tableci{75.4}{1.4} & \tableci{64.8}{0.6} & \tableci{45.2}{0.7} \\
\midrule
\multicolumn{7}{l}{\emph{Supervised baselines (re-evaluated)}} \\
RoBERTa-Large & \tableci{95.5}{0.1} & \tableci{77.3}{0.4} & \tableci{65.4}{2.1} & \tableci{47.2}{1.7} & \tableci{63.1}{0.6} & \tableci{71.7}{0.6} \\
RoBERTa-ChatGPT & \textbf{\tableci{100.0}{0.0}} & \tableci{56.7}{0.5} & \tableci{53.5}{2.2} & \tableci{73.5}{1.4} & \tableci{75.8}{0.5} & \tableci{38.6}{0.6} \\
RADAR & \tableci{55.0}{0.4} & \tableci{61.2}{0.5} & \tableci{71.7}{2.1} & \tableci{73.4}{1.9} & \tableci{76.5}{0.6} & \tableci{76.1}{0.6} \\
ModernBERT-Detect & \tableci{98.6}{0.1} & \tableci{97.9}{0.1} & \underline{\tableci{76.9}{1.8}} & \underline{\tableci{94.8}{0.8}} & \underline{\tableci{87.9}{0.4}} & \underline{\tableci{99.7}{0.1}} \\
Human-as-Outlier & \tableci{99.3}{0.1} & \underline{\tableci{98.0}{0.1}} & \tableci{76.1}{1.8} & \tableci{94.0}{0.8} & \tableci{87.7}{0.4} & \tableci{65.8}{0.6} \\
RepreGuard & \tableci{95.2}{0.1} & \tableci{47.3}{0.5} & \tableci{75.9}{1.8} & \tableci{89.0}{1.2} & \tableci{69.4}{0.6} & \tableci{93.9}{0.3} \\
\midrule
\textbf{MELD} (ours) & \underline{\tableci{99.90}{0.01}} & \textbf{\tableci{99.1}{0.1}} & \textbf{\tableci{77.5}{1.3}} & \textbf{\tableci{99.4}{0.1}} & \textbf{\tableci{99.8}{0.0}} & \textbf{\tableci{99.90}{0.04}} \\
\bottomrule
\end{tabular}

%% file: figures/tab_meld_eval.tex
\begin{tabular*}{\textwidth}{@{\extracolsep{\fill}}lrrrrr@{}}
\toprule
\textbf{Detector} & \textbf{GPT-5.4-Mini} & \textbf{Gemini-3-Flash} & \textbf{Claude-Haiku-4.5} & \textbf{Qwen-3.6-Plus} & \textbf{Overall} \\
\midrule
\multicolumn{6}{l}{\emph{Zero-shot reference-LM detectors}} \\
GLTR & \tableci{1.2}{0.3} & \tableci{0.5}{0.3} & \tableci{1.3}{0.4} & \tableci{0.3}{0.1} & \tableci{0.8}{0.3} \\
Fast-DetectGPT & \tableci{9.2}{1.7} & \tableci{19.8}{2.1} & \tableci{14.3}{2.0} & \tableci{24.7}{2.3} & \tableci{17.0}{2.0} \\
Binoculars & \tableci{0.2}{0.1} & \tableci{0.6}{0.1} & \tableci{1.1}{0.2} & \tableci{0.7}{0.2} & \tableci{0.6}{0.1} \\
\midrule
\multicolumn{6}{l}{\emph{Supervised baselines (re-evaluated)}} \\
RoBERTa-Large & \tableci{1.1}{0.1} & \tableci{1.1}{0.1} & \tableci{1.8}{0.1} & \tableci{1.8}{0.1} & \tableci{1.4}{0.0} \\
RoBERTa-ChatGPT & \tableci{0.5}{0.1} & \tableci{1.2}{0.2} & \tableci{0.5}{0.2} & \tableci{0.9}{0.2} & \tableci{0.8}{0.2} \\
RADAR & \tableci{0.4}{0.2} & \tableci{1.4}{0.5} & \tableci{2.2}{0.8} & \tableci{1.2}{0.5} & \tableci{1.3}{0.5} \\
Human-as-Outlier & \tableci{2.7}{1.2} & \tableci{1.6}{0.9} & \tableci{0.8}{0.4} & \tableci{1.1}{0.8} & \tableci{1.6}{0.8} \\
ModernBERT-Detect & \underline{\tableci{97.2}{1.0}} & \underline{\tableci{93.0}{2.2}} & \underline{\tableci{98.2}{0.9}} & \underline{\tableci{93.7}{2.0}} & \underline{\tableci{95.5}{1.5}} \\
RepreGuard & \tableci{23.0}{1.3} & \tableci{27.2}{1.4} & \tableci{41.2}{1.8} & \tableci{45.4}{2.0} & \tableci{34.2}{1.6} \\
\midrule
\textbf{MELD} (ours) & \textbf{\tableci{99.3}{0.3}} & \textbf{\tableci{99.8}{0.4}} & \textbf{\tableci{99.7}{0.5}} & \textbf{\tableci{100.0}{0.2}} & \textbf{\tableci{99.7}{0.3}} \\
\bottomrule
\end{tabular*}

%% file: figures/tab_qualitative.tex
{\small
\setlength{\tabcolsep}{2pt}
\newcommand{\qualsrc}[2]{\parbox[t]{0.125\textwidth}{\raggedright #1\\ #2}}
\newcommand{\qualsrcdetail}[3]{\parbox[t]{0.125\textwidth}{\raggedright #1\\ #2\\ (#3)}}
\begin{longtable}{@{}r@{\hspace{4pt}}>{\raggedright\arraybackslash}p{0.125\textwidth}>{\raggedright\arraybackslash}p{0.35\textwidth}cccc@{}}
\caption{\textbf{Per-text disagreements on HC3, DetectRL, and MELD-eval.}
Each cell reports the standardized margin of a detector score from the pool-specific FPR=$1\%$ threshold:
$\Phi^{-1}(\text{percentile-in-humans}) - \Phi^{-1}(0.99)$, in standard-deviation units.
The threshold is $0$ by construction, so $\checkmark$ marks correct classification
(AI: margin $>0$; human: margin $\leq 0$) and \xmark{} marks an operating-point error.
A score above every pool human saturates the transform, so $+1.51$ is the maximum expressible margin on MELD-eval.
Rows $1$--$4$ are HC3 clean examples (2 AI, 2 human); rows $5$--$8$ are DetectRL attacked examples (3 AI, 1 human);
rows $9$--$12$ are clean MELD-eval AI examples, one per generator; rows $13$--$16$ are attack-augmented MELD-eval AI examples, one per generator. These disagreement examples show cases where MELD remains on the correct side of the low-FPR threshold while other open-source detectors fail.
\label{tab:qualitative}}\\
\toprule
\# & \textbf{Source / attack} & \textbf{Excerpt} (truncated) & \textbf{MELD} & \textbf{Bino} & \textbf{MBERT-D} & \textbf{RoB-CGPT} \\
\midrule
\endfirsthead

\multicolumn{7}{l}{\normalsize\textit{Table~\ref{tab:qualitative} (continued).}} \\
\toprule
\# & \textbf{Source / attack} & \textbf{Excerpt} (truncated) & \textbf{MELD} & \textbf{Bino} & \textbf{MBERT-D} & \textbf{RoB-CGPT} \\
\midrule
\endhead

\midrule
\multicolumn{7}{r}{\normalsize\textit{continued on next page}} \\
\endfoot

\bottomrule
\endlastfoot

\multicolumn{7}{l}{\emph{HC3 (clean text)}} \\
(1) & \qualsrc{HC3}{clean} & \textit{AI}: John Adams became President in 1797. He was the second President of the United States, serving one term from 1797 to 1801. & \textbf{$+$0.71}\,$\checkmark$ & $-$1.32\,\xmark{} & $-$3.37\,\xmark{} & $+$0.87\,$\checkmark$ \\
\addlinespace[2pt]
(2) & \qualsrc{HC3}{clean} & \textit{AI}: Charcoal and regular coal are similar in that they are both made from carbon-rich materials and they can be used as fuels. However, there are some differences between the two that make burning charcoal less practical... & \textbf{$+$0.58}\,$\checkmark$ & $-$1.52\,\xmark{} & $+$0.42\,$\checkmark$ & $-$6.60\,\xmark{} \\
\addlinespace[2pt]
(3) & \qualsrc{HC3}{clean} & \textit{Human}: * " I 'd imagine it has something to do with availability . " * True . We 'd probably have problems locking up women in cages too , though . & \textbf{$-$1.43}\,$\checkmark$ & $-$0.65\,$\checkmark$ & $+$1.70\,\xmark{} & $-$1.15\,$\checkmark$ \\
\addlinespace[2pt]
(4) & \qualsrc{HC3}{clean} & \textit{Human}: Logical consequence (also entailment) is a fundamental concept in logic, which describes the relationship between statements that hold true when one statement logically follows from one or more statements. A valid logical argument... & \textbf{$-$4.35}\,$\checkmark$ & $-$3.69\,$\checkmark$ & $+$0.37\,\xmark{} & $-$0.58\,$\checkmark$ \\
\midrule
\multicolumn{7}{l}{\emph{DetectRL (attacked)}} \\
(5) & \qualsrc{DetectRL}{paraphrase} & \textit{AI}: Counting them up revealed steady progress one line at a time. The story was gradually taking shape sentence by sentence. Introducing details like checking my count kept the writing engaging. Finding creative ways to el... & \textbf{$+$0.08}\,$\checkmark$ & $-$1.31\,\xmark{} & $-$2.69\,\xmark{} & $-$2.13\,\xmark{} \\
\addlinespace[2pt]
(6) & \qualsrc{DetectRL}{perturbation} & \textit{AI}: I've also noticed that it has a peasant scent. It' not too strong or overpowering, but it's definitely noticeable. It's a nica bonus, especially since I enjoy using products that have a nice fragrance. The packagign i... & \textbf{$+$0.36}\,$\checkmark$ & $-$0.94\,\xmark{} & $-$4.91\,\xmark{} & $-$1.38\,\xmark{} \\
\addlinespace[2pt]
(7) & \qualsrc{DetectRL}{prompt} & \textit{AI}: The only thing that was edible was the soup. The service was also terrible. The waiter was rude and dismissive. He took our order and then disappeared for over 30 minutes. When he finally came back with our food, it w... & \textbf{$+$0.04}\,$\checkmark$ & $-$0.85\,\xmark{} & $-$2.05\,\xmark{} & $-$0.75\,\xmark{} \\
\addlinespace[2pt]
(8) & \qualsrc{DetectRL}{paraphrase} & \textit{Human}: In the hushed living room, my family surrounds me. The sole sound is a news reporter's voice droning from the TV. Dim light bathes the room in an eerie, pulsating blue glow that mimics the rhythm of my heartbeat. On t... & \textbf{$-$0.88}\,$\checkmark$ & $-$1.19\,$\checkmark$ & $+$0.46\,\xmark{} & $-$1.03\,$\checkmark$ \\
\midrule
\multicolumn{7}{l}{\emph{MELD-eval clean}} \\*
(9) & \qualsrcdetail{MELD-eval}{clean}{GPT-5.4-Mini, abstracts} & \textit{AI}: Whitney's broken-circuit theorem gives a foundational expansion of the chromatic polynomial in terms of acyclic substructures of a graph and has played a central role in the combinatorial theory of graph colorings. In this paper... & \textbf{$+$1.23}\,$\checkmark$ & $-$2.93\,\xmark{} & $-$2.10\,\xmark{} & $-$3.10\,\xmark{} \\
\addlinespace[2pt]
(10) & \qualsrcdetail{MELD-eval}{clean}{Gemini-3-Flash, abstracts} & \textit{AI}: The deployment of autonomous robotic systems in dynamic, unstructured environments necessitates perception pipelines capable of high-fidelity spatial reasoning under stringent temporal constraints. Reliable image segmentation... & \textbf{$+$0.48}\,$\checkmark$ & $-$3.17\,\xmark{} & $-$0.56\,\xmark{} & $-$3.79\,\xmark{} \\
\addlinespace[2pt]
(11) & \qualsrcdetail{MELD-eval}{clean}{Claude-Haiku-4.5, abstracts} & \textit{AI}: The Riemann Hypothesis remains one of mathematics' most profound unsolved problems, asserting that all non-trivial zeros of the Riemann zeta function lie on the critical line Re(s) = 1/2. This review synthesizes... & \textbf{$+$0.60}\,$\checkmark$ & $-$2.10\,\xmark{} & $-$1.50\,\xmark{} & $-$2.59\,\xmark{} \\
\addlinespace[2pt]
(12) & \qualsrcdetail{MELD-eval}{clean}{Qwen-3.6-Plus, abstracts} & \textit{AI}: Boolean satisfiability solving has evolved from a theoretically intractable problem into a cornerstone of modern electronic design automation, formal verification, and artificial intelligence. While early algorithms struggled... & \textbf{$+$0.87}\,$\checkmark$ & $-$3.19\,\xmark{} & $-$1.75\,\xmark{} & $-$2.71\,\xmark{} \\
\midrule
\multicolumn{7}{l}{\emph{MELD-eval attacked}} \\*
(13) & \qualsrcdetail{MELD-eval}{upper-lower}{GPT-5.4-Mini, reddit} & \textit{AI}: their gc and thEn whEn i asked whY i coUldn't join they Just Said ``it's complicatEd'' like Okay?? apparently it's fine for literallY everyone else except me. one of them wIll poSt pics of them all hanging out and i'm sit... & \textbf{$+$1.51}\,$\checkmark$ & $-$2.65\,\xmark{} & $-$1.68\,\xmark{} & $-$3.28\,\xmark{} \\
\addlinespace[2pt]
(14) & \qualsrcdetail{MELD-eval}{upper-lower}{Gemini-3-Flash, poetry} & \textit{AI}: The clock face blurs in the heavy silence of Three a.m. The walls are thin, the shadows long, and I am the only pulse in tHis Room. I tiLt my chin toward the wiNdow, searching the blacK velvet of the void for a sign... & \textbf{$+$1.23}\,$\checkmark$ & $-$1.67\,\xmark{} & $-$1.51\,\xmark{} & $-$2.59\,\xmark{} \\
\addlinespace[2pt]
(15) & \qualsrcdetail{MELD-eval}{homoglyph}{Claude-Haiku-4.5, abstracts} & \textit{AI}: Image-to-image translation, the tаsk of converting images frоm one domain to another while preserving content structure, has become increasingly important for applications ranging from style trаnsfer to medical imaging. Recent advances in generative adversarial networks... & \textbf{$+$1.51}\,$\checkmark$ & $-$3.71\,\xmark{} & $-$0.94\,\xmark{} & $-$3.69\,\xmark{} \\
\addlinespace[2pt]
(16) & \qualsrcdetail{MELD-eval}{synonym}{Qwen-3.6-Plus, reddit} & \textit{AI}: I know moving is trying for everyone, but I didn't realize it would hit her this hard. She has always been a pretty chill cat, the kind that sleeps through vacuuming and doesn't even flinch when the doorbell rings. See... & \textbf{$+$1.51}\,$\checkmark$ & $-$2.36\,\xmark{} & $-$3.59\,\xmark{} & $-$2.13\,\xmark{} \\

\end{longtable}
}